\documentclass[final,5p,times,twocolumn]{elsarticle}

\usepackage{amssymb}
\usepackage{amsmath}
\usepackage{bm}
\usepackage{algorithm}
\usepackage{algpseudocode}
\usepackage{enumitem}
\usepackage{xcolor}

\begin{document}

\begin{frontmatter}



\title{Humanoid Robot Running Through Random Stepping Stones and Jumping Over Obstacles: Step Adaptation Using Spring-Mass Trajectories}

\author[label1]{Sait Sovukluk}
\author[label2]{Johannes Englsberger}
\author[label1,label2]{Christian Ott}
\affiliation[label1]{organization={Automation and Control Institute (ACIN)},
            addressline={TU Wien},
            postcode={1040},
            state={Vienna},
            country={Austria}}

\affiliation[label2]{organization={Institute of Robotics and Mechatronics},
            addressline={German Aerospace Center (DLR)},
            postcode={82234},
            state={Weßling},
            country={Germany}}



\begin{abstract}
This study proposes a step adaptation framework for running through spring-mass trajectories and deadbeat control gain libraries. It includes four main parts: (1) Automatic spring-mass trajectory library generation; (2) Deadbeat control gain library generation through an actively controlled template model that resembles the whole-body dynamics well; (3) Trajectory selection policy development for step adaptation; (4) Mapping spring-mass trajectories to a humanoid model through a whole-body control (WBC) framework also accounting for closed-kinematic chain systems, self collisions, and reactive limb swinging. We show the inclusiveness and the robustness of the proposed framework through various challenging and agile behaviors such as running through randomly generated stepping stones, jumping over random obstacles, performing slalom motions, changing the running direction suddenly with a random leg, and rejecting significant disturbances and uncertainties through the MuJoCo physics simulator. We also perform additional simulations under a comprehensive set of uncertainties and noise to better justify the proposed method’s robustness against real-world challenges, such as signal noises, imprecision, modeling errors, and delays. All the aforementioned behaviors are performed with a single library and the same set of WBC control parameters without additional tuning. The spring-mass and the deadbeat control gain library are automatically computed in 4.5 seconds in total for 315 different trajectories.
\end{abstract}




\end{frontmatter}



\section{Introduction}
The spring-mass model, also known as the spring-loaded inverted pendulum (SLIP), was first introduced to model the steady-state running of animals and humans \cite{slipref1}, in which running is defined as the center-of-mass (CoM) bouncing on a springy leg through single-leg support and flight phases. Comprehensive experimental and dimensionless mathematical analyses revealed that the ground reaction force pattern of steady-state running of humans and animals, ranging from insects to horses, can be modeled through the spring-mass model, surprisingly, with almost the same relative (dimensionless) leg stiffness ($\mathtt{\sim}$10) \cite{slipref2, slipref3}. It is also revealed that ground reaction force patterns for human walking are also closely related to the spring-mass model \cite{slipWalk}. Even though the spring-mass model is a simplification, it can generate nearly every walking and running gait observed in people and animals \cite{jonathan_article}.

While a simple spring-mass model from a biomechanics perspective, SLIP exhibits many control challenges as it is a hybrid system with nonlinear stance dynamics \cite{deadbeat3}. Seyfarth et al. \cite{criterion_for_running} studied the stability characteristics of planar spring-mass dynamics, and they discovered a J-shaped dependency in the adjustment of the angle of attack to leg stiffness. Similar findings triggered the development of apex-to-apex deadbeat controllers to find out how the leg states (e.g., angle, stiffness, damping) should be modified to maintain stability and convergence during the next stance phase \cite{deadbeat3, deadbeat1, deadbeat2}. The first successful realization of the spring-mass model appeared in Raibert's hoppers, runners, and walkers \cite{raibert1986legged}. ATRIAS is another bipedal robot that is carefully designed to match the spring-mass dynamics and was the first machine to demonstrate humanlike walking-gait dynamics \cite{jonathan_atrias}. Mapping a 3D spring-mass (SLIP) model to a full-scale robotic human model was first studied in \cite{wensingRunning} and later on \cite{sovukluk_slip}. It's achieved via obtaining the desired center of mass (CoM) trajectories through the spring-mass model and then commanding them to a robotic human model through a whole-body controller (WBC) while keeping the torso upright and employing proper leg and arm swinging. However, both \cite{wensingRunning} and \cite{sovukluk_slip} lack step adaptation and focus only on velocity control. Furthermore, as the spring-mass model does not account for articulated body dynamics, mapping such center of mass trajectories to a full-scale humanoid robot requires careful tuning of limb-swing trajectories to ensure that the articulated body and angular momentum effects are properly handled at the whole-body control level. A different center of mass trajectory may require a different limb-swing behavior due to changes in the force profiles, velocities, stance time, and flight time, preventing the utilization of various CoM trajectories for step adaptation purposes through the same limb-swing control. In this study, we address both the step adaptation and reactive limb-swing trajectory generation problems, enabling navigation through random environments through switching between various CoM trajectories without any trajectory-specific tuning. The framework's agility and disturbance rejection make it suitable for outdoor challenges, such as search-and-rescue or industrial inspection in unstructured environments.

The literature on dynamic humanoid running is largely lacking in step adaptation and appears more limited compared to studies on walking. Furthermore, different from walking, where there is always contact with the ground, running is more demanding in terms of power outputs, peak forces, impacts, state estimation, and contact conditions as the robotic system alternates between the stance and flight phases on a single leg, requiring exceptional hardware that is also precise and impact-resistant. Consequently, most of the full-scale humanoid running literature appears in simulation environments \cite{walkingToRunningRome, kuindersma2016optimization, dai2014, HzdRun, recent_run_2, recent_run_3, recent_run_5}, and the experimental verifications are usually limited to velocity-based running with insignificant flight phases and without step adaptation \cite{runFast, asimoRun, recent_run_1, recent_run_4}. The recent machine learning literature reveals that reinforcement learning-based methods for running are mostly based on imitation learning, which requires a strong reference from human running \cite{LocoMuJoCo, protomotions, recent_run_1}. As imitation is a mapping from human data, the generalization of running behaviors for random step adaptation has not yet been demonstrated.

Dynamic humanoid running with step adaptation and disturbance rejection capabilities without prior environmental knowledge is an understudied topic that has not been thoroughly addressed in the literature. The combination of the nonlinear and hybrid nature of running dynamics with foothold constraints presents a challenging problem that requires special care. This problem is easier and has already been addressed in the case of walking \cite{ihmc_robert}. Walking dynamics is more flexible, stable, and easier to solve, and also allows static walking to preserve stability through balancing, as there is always at least one leg on the ground. There are two running frameworks appearing with step adaptation capability: BID \cite{BID} and DCM-based running \cite{DCM}. Different from the spring-mass model, where the dynamics are nonlinear and not closed-form solvable, both methods provide closed-form solvable linear running dynamics with a trade-off of preselection requirement of some stability- and feasibility-critical parameters, including but not limited to: stance time, flight time, and touchdown height, along with the desired foot placement point. Furthermore, improper stance time selection results in leg overextension as the CoM moves far away from the foot placement point during the stance phase. Similarly, the touchdown height selection also affects stability and may result in violating the friction constraint. Hence, the feasibility and stability in both studies depend on manual parameter selections, requiring a trial-and-error process for different environments, running velocities, and types. BID-based planning includes a disturbance rejection controller, whereas DCM-based planning does not. The proposed step adaptation framework, on the other hand, is tuning-free and only requires selection for the desired range for running velocity, apex height, and leg stiffness. Compared to \cite{BID}'s 2016 BID and \cite{DCM}'s 2023 DCM running, our work eliminates manual parameter tuning for stance/flight times and determines the stability critical parameters inherently, also considering feasibility, stability, disturbances, errors, and desired foot placement points. As a result, the proposed step adaptation framework with reactive limb-swing capabilities enables real-time adaptation to random stepping stones without prior environmental information.
\begin{figure}[t!]
\centerline{\includegraphics[width=0.90\columnwidth]{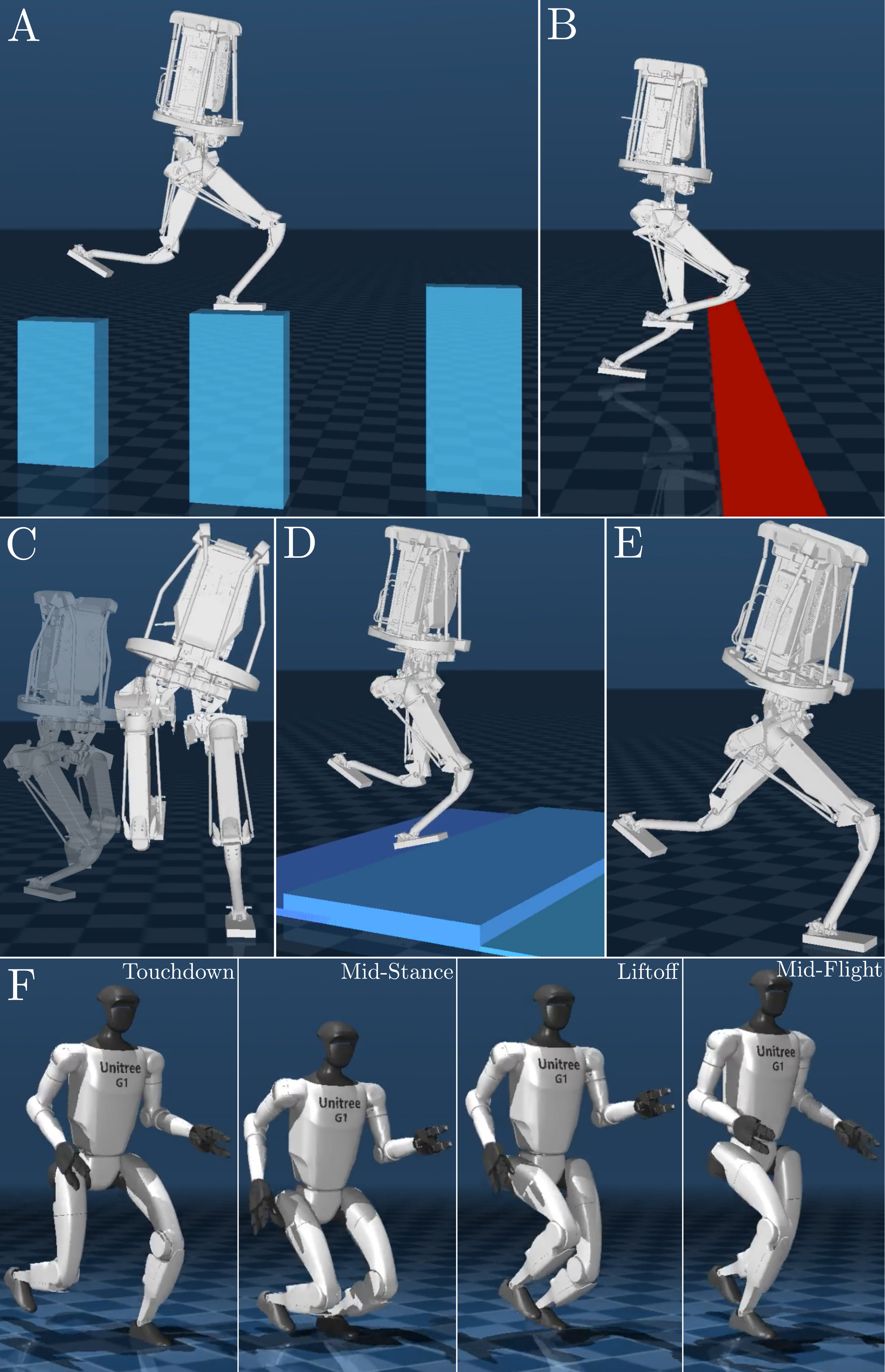}}
\caption{(A) Running through randomly generated stepping stones; (B) Jumping over randomly generated obstacles; (C) Sudden direction change with the inner leg; (D) Running through unobserved height differences; (E) Rejecting a forward push disturbance with a wide step; (F) Snapshots of Unitree G1 running. Multiple robots are presented to show the generalizability of the proposed framework. The supplemental video collects all and more.}
\label{firstPageFig}
\end{figure}
\begin{figure*} [t!]
\centerline{\includegraphics[width=\textwidth]{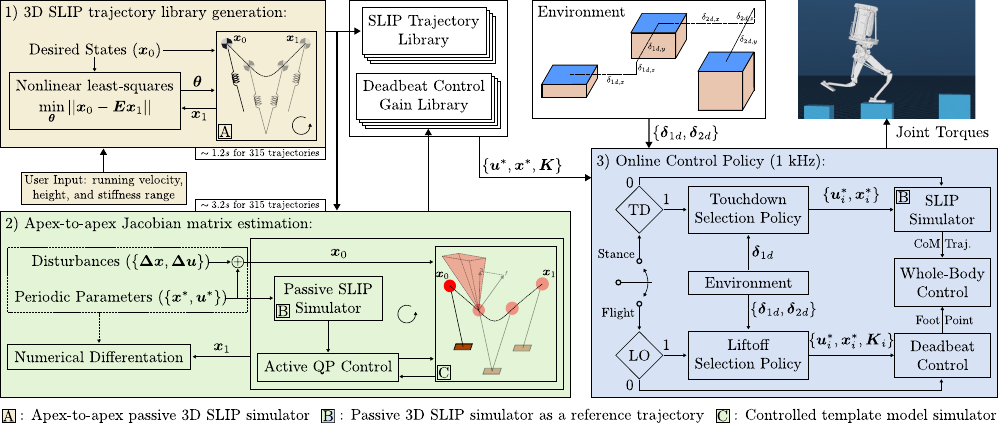}}
\caption{Overview of the proposed framework. Three main steps include (1) 3D SLIP periodic trajectory search, (2) apex-to-apex Jacobian estimations through an actively controlled system that resembles the whole-body dynamics well, and (3) online control that combines trajectory selection policies and the whole-body controller with a reactive limb swing task.}
\label{overviewFig}
\end{figure*}
\subsection{Contribution}
\begin{itemize}
    \item The spring-mass model is a velocity-based running model, i.e., the error is defined in terms of velocity and jumping height, and is constructed around periodic trajectories. Hence, achieving step adaptation to run through random stepping stones is the main contribution of this paper. The step adaptation is achieved through switching between different trajectories contained in an automatically generated trajectory library, which takes $\sim$4.5 seconds to generate for 315 different trajectories. We first develop the necessary conditions for convergence between periodic trajectories and analyze the minimum number of steps required for planning. Then, develop a trajectory selection policy that switches between adequate trajectories depending on the foot placement point requirements, considering stability, kinematic and dynamic feasibility, and the convergence conditions. The proposed method is carefree, and the stability critical parameters, such as stance time, flight time, and touchdown angle, are determined inherently.
    \item The CoM trajectories, that are obtained from reduced or template models, are realized on full-scale humanoid robots through whole-body controllers as a CoM trajectory tracking task combined with other tasks such as torso orientation, leg swing, and arm swing control through hand-crafted limb swing trajectories \cite{wensingRunning, sovukluk_slip, BID, DCM}. However, in the case of frequent trajectory switching and disturbances, where the force profiles, stance time, flight time, and foot placement conditions vary, a different limb-swing behavior is required to maintain postural stability. We address this problem through a reactive limb swing control task that utilizes dynamic coupling between the links and the base frame orientation, thereby enhancing disturbance rejection capabilities. We utilize the redundant leg and arm joints and control their states depending on the base frame orientation error, such that the postural stability is preserved while the robot follows the given CoM trajectories and foot placement point commands. The extensive set of behaviors proposed in this study (also shown in Fig.~\ref{firstPageFig} and the supplemental video) is realized using a single library and the same whole-body controller, without any additional parameter or trajectory tuning.
    \item This work combines a diverse set of dynamic behaviors in a single framework. These behaviors include running through previously unknown, randomly generated stepping stones, jumping over suddenly appearing obstacles, sudden turns and slalom motions, and disturbance rejection, including force disturbances, unobserved height differences, and edge stepping. To further justify the proposed method's robustness against real-world challenges, we also simulate its operation under a comprehensive and significant set of uncertainties and noise, including signal noise, modeling errors, imprecision, and delays.
\end{itemize}
\subsection{Content}
The flow of the paper is summarized in Fig.~\ref{overviewFig}. The paper starts with a summary of the spring-mass dynamics and periodic trajectory optimization formulation. Then, it proceeds with trajectory library generation, observations on trajectory characteristics, and calculation of stepping distances. The development continues with the computation of deadbeat control gain through an actively controlled spring-mass model for stabilization, disturbance rejection, and trajectory switching purposes. The third part is related to the development of the step adaptation policy. We first analyze how many steps to look ahead to guarantee convergence in the spring-mass dynamics level. We then propose two sets of algorithms to jump over obstacles and run through stepping stones considering feasibility, stability, and convergence conditions. The fourth part includes conventional inverse-dynamics whole-body control formulations with two add-ons: a task formulation for reactive limb swing and self-collision avoidance. Then, we discuss computation times and simulation results for an extensive range of behaviors, including noise and uncertainty analysis. Lastly, a conclusion section relates the paper's objective to its methodology and results, and summarizes the outcomes.
\section{3D Spring-Mass Trajectory Library}
This section involves constructing a nonlinear least-squares problem to solve for periodic spring-mass trajectories. Then, this structure is used to generate a trajectory library that combines different types of running in terms of apex height, velocity, stiffness, and lateral leg angle. This section corresponds to the first block of Fig.~\ref{overviewFig} and discusses mapping spring-mass trajectories to foot placement point information.
\subsection{Background: 3D Spring-Mass Dynamics}
\begin{figure}[t!]
\centerline{\includegraphics[width=5cm]{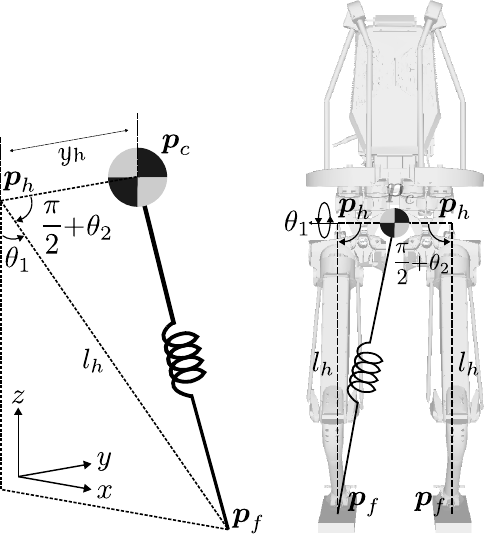}}
\caption{3D SLIP model representation. Parameter $y_{h}$ separates the robot's center of mass and hip positions. The virtual leg has a length of $l_{h}$ and connects the hip to the foot. The touchdown angle is determined by $\theta_{1}$ and $\theta_{2}$. As the virtual leg is offset by $y_{h}$, if $\theta_{2}$ is zero, the foot is in the hip sagittal (parasagittal) plane. Otherwise, the lateral foot placement point is wider or narrower than the virtual hip point $y_{h}$.}
\label{slipModelFig}
\end{figure}

Defining $\bm{p}_{c} \in \mathbb{R} ^ {3}$ to be the position of CoM, the stance dynamics of the point foot 3D spring-mass model (see Fig.~\ref{slipModelFig}) is
\begin{equation} \label{slipStanceDynEq}
    m\ddot{\bm{p}}_{c} = k(r_{0} - ||\bm{r}||) \hat{\bm{r}} + m\bm{g},
\end{equation}
where $m$, $k$, $r_{0}$, and $\bm{g}$ are the total mass, spring constant, leg rest length, and gravitational vector. Additionally, $\bm{r} = \bm{p}_{c} - \bm{p}_{f}$ represents the leg vector, i.e., a vector from the foot point to the point mass (see Fig.~\ref{slipModelFig}) and $\hat{\bm{r}}$ is its unit vector. Through the same notation with \cite{wensingRunning} and \cite{sovukluk_slip}, the virtual hip point is located at $\bm{p}_{h}$ and it helps to decompose leg angles $\theta_{1}$ and $\theta_{2}$ into more intuitive quantities. During the flight phase, the foot point $\bm{p}_{f}$ is a function of $\bm{p}_{c}$, leg angles, hip distance, and leg length,
\begin{equation} \label{slipFootPointEq}
    \bm{p}_{f} = \bm{p}_{c} + 
    \begin{bmatrix}
    0\\
    \sigma y_{h}\\
    0
    \end{bmatrix}
    +
    l_{h}
    \begin{bmatrix}
    \sin{\theta_{1}}\cos{\theta_{2}}\\
    \sigma\sin{\theta_{2}}\\
    -\cos{\theta_{1}}\cos{\theta_{2}}
    \end{bmatrix},
\end{equation}
where $\sigma$ is a sign multiplier and changes between $+1$ and $-1$ depending on which leg is the stance leg. At the moment of touchdown, $\bm{p}_{f}$ is fixed on the ground and sets an initial condition for the stance phase dynamics \eqref{slipStanceDynEq}. A touchdown and liftoff condition separate the stance and flight phase dynamics such that the overall dynamics $\Sigma$ is given by
\begin{equation} \label{slipOverallDynEq}
    \Sigma : 
    \begin{cases}
    \enspace \ddot{\bm{p}}_{c} = \dfrac{k}{m}(r_{0} - ||\bm{r}||) \hat{\bm{r}} + \bm{g} \enspace & ||\bm{r}|| < r_{0} \\
    \enspace \ddot{\bm{p}}_{c} = \bm{g} \enspace & ||\bm{r}|| \geq r_{0}
    \end{cases}.
\end{equation}
\subsection{Background: Apex-to-apex Periodic Trajectory Search}
Studies in \cite{slip_unique1} and \cite{slip_unique2} showed that for a constant energy level of a planar point-mass model, there exists a unique relationship between system parameters. For example, for a specific spring constant, running height, and running velocity, there exists a unique touchdown angle that results in a periodic limit cycle. Consequently, if $\theta_{1}$ is the touchdown angle of a planar point-mass model, this parameter is a function of running height $h$, spring constant $k$, and running velocity $v_{x}$, i.e., $\theta_{1}(h, k, v_{x})$. Selecting $\bm{\theta} = [\theta_{1}, \theta_{2}]^{\top}$ and $\bm{v} = [v_{x}, v_{y}]^{\top}$, one can rewrite the touchdown angle function as $\bm{\theta}(h, k, \bm{v})$.

For an apex state definition of $\bm{x} = [v_{x}, v_{y}, h]^{\top}$ a mapping function from one apex state to the next one can be defined as $P: (\bm{x}_{j}, \bm{\theta}, k) \rightarrow \bm{x}_{j+1}$. Accordingly, for a selection of $h$, $k$, and $\bm{v}$, the periodic trajectory search problem can be formulated as a nonlinear least-squares problem,
\begin{equation} \label{slipSearchEq0}
    \min_{\bm{\theta}} || \bm{x}_{j} - \bm{E}\bm{x}_{j+1} ||,
\end{equation}
where $E = \mathrm{diag}(1,-1,1)$ is placed for sign changes due to the leg switching between two subsequent apex states. Similarly, since it is not intuitive to pick a proper value for $v_{y}$, for a selection of $h$, $k$, $v_{x}$, and $\theta_{2}$, the nonlinear least-squares problem can be reformulated:
\begin{equation} \label{slipSearchEq}
    \min_{\theta_{1}, v_{y}} || \bm{x}_{j} - \bm{E}\bm{x}_{j+1} ||.
\end{equation}
\begin{figure}[t!]
\centerline{\includegraphics[width=5cm]{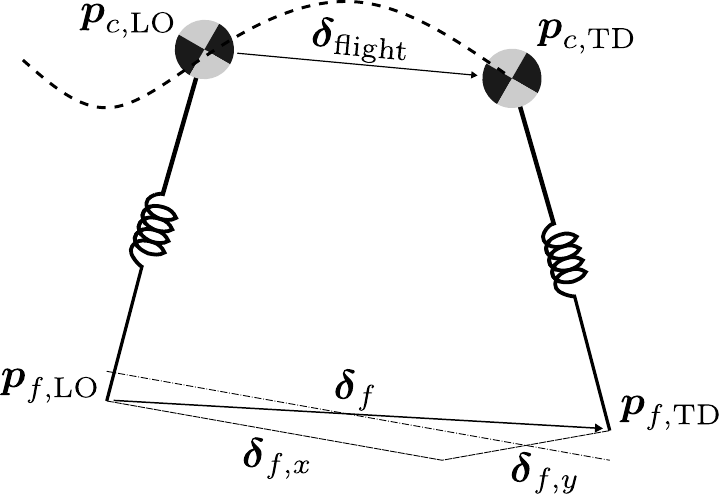}}
\caption{An illustration for foot placement distance between two consecutive steps on a constant height flat surface.}
\label{slipDistanceFig}
\end{figure}
\subsection{Construction and Interpretation of Trajectory Library}
A combination of different types of running can be stored in a trajectory library, allowing the robot to switch between different trajectories and adapt to changing environmental conditions. Then, to achieve step adaptation, the trajectory information must be transformed into a more useful form, namely, stepping distances. The distance between two consecutive foot placement points can be formulated as
\begin{equation} \label{stepLength}
    \bm{\delta}_{f} = \underbrace{(\bm{p}_{c,\text{LO}} - \bm{p}_{f,\text{LO}})}_{\bm{r}_{\text{LO}}} + \bm{\delta}_{\text{flight}} + \underbrace{(\bm{p}_{f,\text{TD}} - \bm{p}_{c,\text{TD}})}_{-\bm{r}_{\text{TD}}},
\end{equation}
where $(\cdot)_{\text{TD}}$ and $(\cdot)_{\text{LO}}$ represent parameters at the moment of touchdown and lift-off, respectively. Additionally, $\bm{\delta}_{\text{flight}}$ represents the CoM displacement during the flight phase (see Fig.~\ref{slipDistanceFig}). As a result, a library $\mathcal{T} \in \{\mathbb{R}^{1\times 6}, \mathbb{S}, \mathbb{S}\}^{n}$ with $n$ trajectories forms the following table:
\begin{equation*} \label{trajectoryTable}
\begin{tabular}{ccccccccc}
 $\theta_{2}$ & $h_{\text{apex}}$ & $k$ & $v_{x,\text{apex}}$ & $\theta_{1}$ & $v_{y,\text{apex}}$ & $\delta_{f,x}$ & $\delta_{f,y}$ \\ \hline
 $\vdots$ & $\vdots$ & $\vdots$ & $\vdots$ & $\vdots$ & $\vdots$ & $\vdots$ & $\vdots$
\end{tabular}
\end{equation*}
where $\theta_{2}$, $h_{\text{apex}}$, $k$ and $v_{x,\text{apex}}$ are selected/desired parameters; $\theta_{1}$ and $v_{y,\text{apex}}$ are result of periodic trajectory search problem \eqref{slipSearchEq}; $\delta_{f,x}$ and $\delta_{f,y}$ are result of \eqref{stepLength}. Note that $\delta_{f,x}$ and $\delta_{f,y}$ change as the flight time, surface height, leg length, and touchdown angle change. The deadbeat controller, which is proposed in the following sections, may also modify the touchdown angle and leg length for disturbance rejection purposes. Hence, $\delta_{f,x}$ and $\delta_{f,y}$ should be recalculated at touchdown and liftoff moments.
\begin{figure}[t!]
\centerline{\includegraphics[width=0.75\columnwidth]{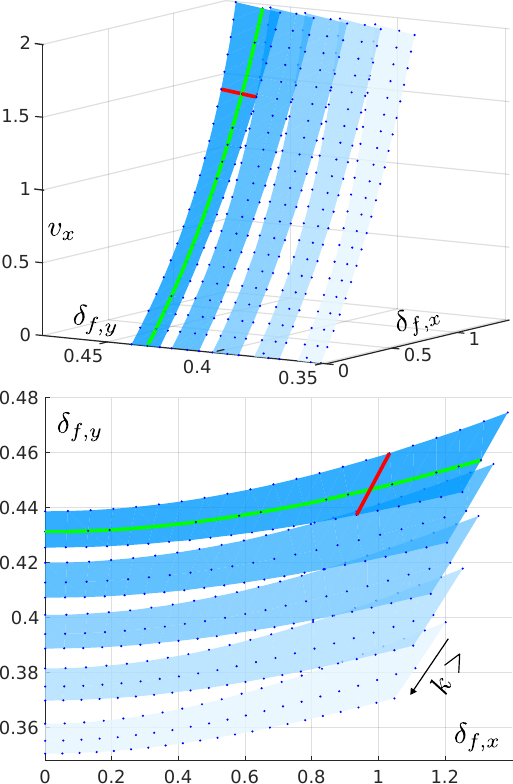}}
\caption{An example trajectory library for undisturbed periodic flat-surface running with constant $\theta_{2}$ (top) and its projection onto the stepping distance $(\delta_{f,x}, \delta_{f,y})$ plane as shown in Fig.~\ref{slipDistanceFig} (bottom). In the library, the running velocity $v_{x}$, leg stiffness $k$, and apex height $h_\text{apex}$ vary. Each tone of blue is a collection of a constant apex height, and darker colors represent higher values. The green line is an example set of trajectories with constant stiffness and varying forward velocity. Similarly, the red line is an example set of trajectories with constant forward velocity but varying leg stiffness. Lastly, the black arrow represents the direction of increasing leg stiffness for each constant apex height surface.}
\label{libraryFig}
\end{figure}
\begin{figure}[t!]
\centerline{\includegraphics[width=0.75\columnwidth]{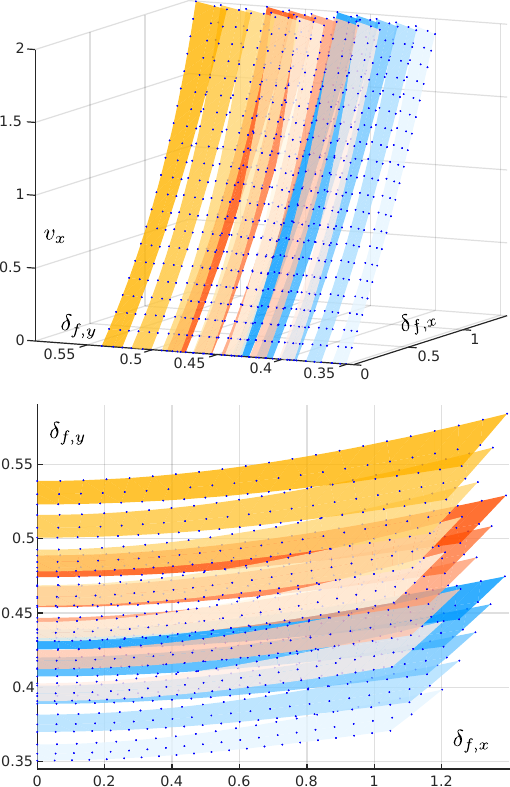}}
\caption{Extended trajectory library over Fig.~\ref{libraryFig} with varying lateral leg angle $\theta_{2}$. Blue, orange, and yellow represent increasing $\theta_{2}$ in the same order.}
\label{libraryFullFig}
\end{figure}

The foot placement estimation strategy differs in the touchdown and liftoff moments. On the other hand, in the case of an undisturbed periodic running, both estimations output the same result. Assume $\theta_{i,1}^{*}$ and $\theta_{i,2}^{*}$ are the periodic longitudinal and lateral leg angle of the $i^{th}$ trajectory. As a result of the symmetricity of periodic SLIP trajectories, $\theta_{1,\text{LO}} = -\theta_{i,1}^{*}$ and $\theta_{2,\text{LO}} = \theta_{i,2}^{*}$. Hence, substitution of \eqref{slipFootPointEq} into \eqref{stepLength} leads to
\begin{equation}
    \bm{\delta}_{f,i} = \bm{\delta}_{\text{flight}} + \underbrace{2 
    \begin{bmatrix}
    0\\
    \sigma y_{h}\\
    0
    \end{bmatrix}
    +
    2l_{h}
    \begin{bmatrix}
    \sin{(-\theta_{i,1}^{*})}\cos{\theta_{i,2}^{*}}\\
    \sigma\sin{\theta_{i,2}^{*}}\\
    -\cos{(-\theta_{i,1}^{*})}\cos{\theta_{i,2}^{*}}
    \end{bmatrix}}_{\bm{r}_{\text{LO}} - \bm{r}_{\text{TD}}}.
\end{equation}
Accounting for possible height changes, define total flight time $t_{\text{flight}} = t_{\text{rise}} + t_{\text{fall}}$ and the foot height w.r.t. ground at the apex state $p_{f,z,\text{apex}} = h^{*} - l_{h}\cos{\theta_{1}^{*}}\cos{\theta_{2}^{*}}$. As a result, $t_{\text{rise}} = \sqrt{2 p_{f,z,\text{apex}}/g}$. Similarly, for a height difference $\delta_{z}$ between the consecutive steps, $t_{\text{fall}} = \sqrt{2(p_{f,z,\text{apex}} - \delta_{z})/g}$ such that the fall time is longer or shorter depending on $\delta_{z}$. Consequently,
\begin{equation}
    \bm{\delta}_{\text{flight}} = t_{\text{flight}}
    \begin{bmatrix}
        v_{x}^{*} \\
        v_{y}^{*} \\
        0
    \end{bmatrix}
    +
    \begin{bmatrix}
        0 \\
        0 \\
        \delta_{z}
    \end{bmatrix}.
\end{equation}

Note that in case of remaining error at the liftoff moment, the system is not periodic anymore, and the symmetricity does not hold. At the liftoff moment, $(\bm{p}_{c,\text{LO}} - \bm{p}_{f,\text{LO}})$ can be calculated through the system states. Leg vector for the next touchdown $(\bm{p}_{f,\text{TD}} - \bm{p}_{c,\text{TD}})$, on the other hand, should be calculated accounting for the deadbeat controlled leg states.
\subsubsection{Visualization of an Example Trajectory Library}
Visualization of an example periodic trajectory library for a constant lateral leg angle $\theta_{2}$ is shown in Fig.~\ref{libraryFig}. In the library, the parameter variations are as follows: $v_{x} = \{0,0.1,\dots,2\}m/s$, $h = \{0.90, 0.925, 0.95, 0.975, 1\}m$, $k = \{6000, 8000, 10000\} N/m$. The parameter ranges are determined considering the hardware limits and kinematics of Kangaroo. The figure shows that running with a constant apex height forms a surface relative to the remaining parameters: running velocity and stiffness. As the flight time is closely related to the apex height, for the same running velocity, the stepping distance increases proportionally with the apex height. Furthermore, as decreased leg stiffness results in more inclined touchdown and lift-off angles, foot placement distance increases as the stiffness decreases for the same running velocity. Each set of points $(\delta_{f,x}, \delta_{f,y}, v_{x})$ is unique in 3D space, and their projection onto the stepping distance plane may overlap. The variation of $\theta_{2}$ introduces additional depth in 3D space and new points on the $\delta_{f,x}\delta_{f,y}$ plane (see Fig.~\ref{libraryFullFig}). The most observable characteristic of points with different lateral leg angles is the overall shift in the $\delta_{f,y}$ axis, i.e., the lateral displacement $\delta_{f,y}$ increases or decreases proportionally with $\theta_{2}$. All behaviors shown in this work are generated using the same library shown in Fig.~\ref{libraryFig}.
\section{Deadbeat Control of Active Spring-Mass Model}
\begin{figure}[t!]
\centerline{\includegraphics[width=0.5\columnwidth]{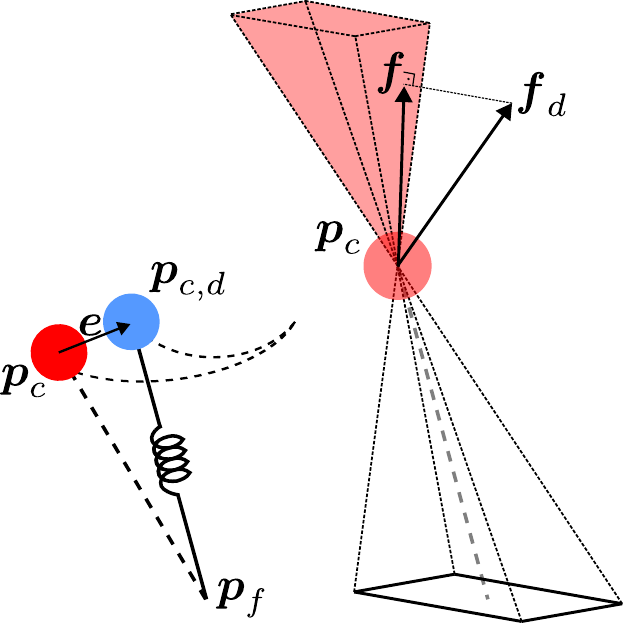}}
\caption{Left: The error definition of the control system where the blue SLIP represents the desired undisturbed periodic state evolution and the red is the actively controlled system. Right: A detailed sketch of the actively controlled template model with a foot. If the desired total control force $\bm{f}_{d}$ is outside the feasible polyhedral cone, it is projected to the closest feasible solution.}
\label{slipFootFig}
\end{figure}

Running is a combination of two consecutive phases: the single support stance and the flight phase. As the floating base dynamics is driven by the ground reaction forces \cite{holonomy, CMM1, CMM2, wensing2023optimization}, they play a vital role in locomotion. During the stance phase, the robotic system drives its joints and the ground reaction forces such that the CoM and limbs track the desired trajectories. The flight phase is fully ballistic and can be considered as a posture preparation for the following stance phase. As legged robots are not fixed but have floating base systems, the set of feasible ground reaction forces is strictly constrained by foothold, friction, and positive vertical force constraints to prevent leg tilting, sliding, and leg separation, respectively. Since the foot region limits the set of feasible ground reaction forces (see Fig.~\ref{slipFootFig}) and the foot location is kept constant during the stance phase, the relative foot placement point is crucial and determines an initial condition for the stance phase dynamics. The foot should be placed at a point such that the resultant feasible polyhedral cone collects the set of ground reaction forces that are needed to converge to the desired trajectory or to reject disturbances.

This section aims to obtain deadbeat control gains through a model that resembles the full-order whole-body controlled system's CoM dynamics well in terms of control force feasibility. For this purpose, instead of conventional point-foot passive 3D spring-mass dynamics where the leg stiffness is varied for compression and decompression phases for control purposes, this study employs an actively controlled CoM model with a plane foot to estimate each trajectory's apex-to-apex Jacobians and deadbeat control gains. The proposed template model inherits the active control mechanism, friction constraints, and the feasible polyhedral force cone conditions. Consequently, the estimated gains are more relatable to whole-body controlled humanoid models in terms of force and behavioral feasibility. The deadbeat control gain matrix $\bm{K} \in \mathbb{R}^{3 \times 3}$ relates the flight phase state errors to the leg angle and length offsets such that the feasible set of ground reaction forces results in eliminating errors or converging into a new trajectory during the next stance phase. The proposed model assumes that the centroidal angular momentum and additional limb controls are handled at the whole-body control level, which will be discussed in detail in the following sections.
\subsection{Point Mass Model and Control Force}
Let
\begin{equation} \label{slipStanceFootDynEq}
    m\ddot{\bm{p}}_{c} = m\bm{g} + \bm{f},
\end{equation}
be the stance dynamics for point mass dynamics where $\{\bm{f}\in\mathcal{F}^{3} \enspace \vert \enspace \mathcal{F}^{3} \subset \mathbb{R}^{3}\}$ is the set of feasible leg forces. Starting from the initial condition where both systems have the same touchdown point and integrating the undisturbed periodic SLIP dynamics, one can get the desired position, velocity, and acceleration evolution of the CoM dynamics (see Fig.~\ref{slipFootFig}). The difference between the actual and desired trajectories forms the error definition $\bm{e} = \bm{p}_{c,d} - \bm{p}_{c}$. In the case of $\bm{f} = \bm{f}_{d}$, where,
\begin{equation} \label{fdEq}
    \bm{f}_{d} = m\ddot{\bm{p}}_{c,d} - m\bm{g} + \bm{K}_{D}\dot{\bm{e}} + \bm{K}_{P}\bm{e},
\end{equation}
the closed-loop dynamics is exponentially stable:
\begin{equation} \label{optimalCLEq}
    m\ddot{\bm{e}} + \bm{K}_{D}\dot{\bm{e}} + \bm{K}_{P}\bm{e} = \bm{0}.
\end{equation}
In the case the desired control force is not within the feasible set, i.e., $\bm{f}_{d}\notin\mathcal{F}^{3}$, the control force must be projected onto the closest feasible solution. Let $\bm{V} = \begin{bmatrix} \hat{\bm{v}}^{\bm{p}_{c}}_{\bm{p}_{f1}} & \hat{\bm{v}}^{\bm{p}_{c}}_{\bm{p}_{f2}} & \hat{\bm{v}}^{\bm{p}_{c}}_{\bm{p}_{f3}} & \hat{\bm{v}}^{\bm{p}_{c}}_{\bm{p}_{f4}} \end{bmatrix} \in \mathbb{R}^{3 \times 4}$ represents the combination of all corner unit vectors from foot corners to the CoM. A linear combination of the corner vectors represents all feasible forces within the foot region. Consequently, for a set of scalar multipliers $\bm{\alpha} \in \mathbb{R}^{4}$, the overall control force formulation with the feasibility constraints takes the following form:
\begin{subequations} \label{QPeqSLIP}
    \begin{equation} \tag{\ref{QPeqSLIP}}
    \underset{\bm{\alpha}}{min} \enspace {|| \bm{f}_{d} - \bm{V}\bm{\alpha} ||}
    \end{equation}
    \begin{equation*}
        \text{Such that:}
    \end{equation*}
    \begin{equation} \label{sc1}
    \bm{\alpha} \geq 0,
    \end{equation}
    \begin{equation} \label{sc2}
    \dfrac{|(\bm{V}\bm{\alpha})_{x}|}{(\bm{V}\bm{\alpha})_{z}} < \mu, \quad \dfrac{|(\bm{V}\bm{\alpha})_{y}|}{(\bm{V}\bm{\alpha})_{z}} < \mu.
    \end{equation}
\end{subequations}
The positive ground reaction force and friction cone constraints are handled by \eqref{sc1} and \eqref{sc2}, respectively. If the minimization problem \eqref{QPeqSLIP} converges to zero, then $\bm{f}_{d}$ is already within the polyhedral limits, e.g., $\bm{f} = \bm{V}\bm{\alpha} = \bm{f}_{d}$. Otherwise, $\bm{f} = \bm{V}\bm{\alpha}$ is the closest projection. In the case of projection, the exponential stability is disturbed and convergence is not guaranteed anymore. As the polyhedral cone moves and rotates as $\bm{p}_{f}$ changes (see Fig.~\ref{slipFootFig}), a step adaptation must be performed to keep $\bm{f}_{d}$ inside the feasible set $\mathcal{F}^{3}$ to achieve stability and convergence.
\subsection{Apex to Apex Deadbeat Control Gain Estimation}
Selecting the apex state as the Poincar\'e section, a discrete mapping from one apex to the next one for the actively controlled point mass dynamics \eqref{slipStanceFootDynEq} can be defined as $P_{\text{active}}:~(\bm{x}_{j}, \bm{u}_{j}) \rightarrow \bm{x}_{j+1}$, where $\bm{x}_{j} = [v_{x}, v_{y}, h]^\top$ is apex velocities and height; $\bm{u}_{j} = [\theta_{1}, \theta_{2}, l_{h}]^\top$ is leg angles and virtual leg length (see Fig.~\ref{slipModelFig}). Assuming $(\bm{x}^{*},\bm{u}^{*})$ results in a periodic trajectory and defining $\bm{\Delta x} = \bm{x} - \bm{x}^{*}$ and $\bm{\Delta u} = \bm{u} - \bm{u}^{*}$, the first-order Taylor Series approximation leads to
\begin{equation} \label{slipTaylorEq}
\begin{split}
    \bm{x}_{k+1} &= P_{\text{active}}(\bm{x}_{k}^{*} + \bm{\Delta x}, \bm{u}^{*} + \bm{\Delta u}) \\
    &\approx \bm{E} \bm{x}_{k}^{*} + \bm{J}_{x} \bm{\Delta x} + \bm{J}_{u} \bm{\Delta u},
\end{split}
\end{equation}
where $\bm{J}_{x} = \partial P_{\text{active}}/\partial\bm{x} \in \mathbb{R}^{3 \times 3}$ and $\bm{J}_{u} = \partial P_{\text{active}}/\partial\bm{u} \in \mathbb{R}^{3 \times 3}$ are Jacobians of the return map evaluated at $(\bm{x}^{*},\bm{u}^{*})$ and $\bm{E} = \mathrm{diag}(1,-1,1)$ is placed for sign changes due to the leg switching. Deviations from periodic apex states can be rejected by changing the leg states for the next touchdown:
\begin{equation} \label{slipJacEq}
 \bm{J}_{u} \bm{\Delta u} = -\bm{J}_{x} \bm{\Delta x}.
\end{equation}
One can solve \eqref{slipJacEq} and obtain a gain matrix $\bm{K} \in \mathbb{R}^{3 \times 3}$ that offsets the periodic leg states depending on error:
\begin{equation} \label{slipDeadbeatEq}
    \bm{u} = \bm{u}^{*} + \bm{K} \underbrace{(\bm{x} - \bm{x}^{*})}_{\bm{\Delta x}},
\end{equation}
where $\bm{K} = -\bm{J}_{u}^{-1}\bm{J}_{x}$. A deadbeat control gain library $\mathcal{K}$ collects gains for each trajectory such that $\mathcal{K} = \{\bm{K}_{0}, \dots, \bm{K}_{n}\}$.
\section{Step Adaptation}
The spring-mass model is a velocity-based running model, constructed around periodic trajectories, and is stabilized through control of leg angle and length at touchdown. Consequently, if no step adaptation is necessary, the method does not require any environmental information beyond touchdown detection, which is necessary for switching from the flight phase to the stance phase. On the other hand, running through discrete surfaces, such as stepping stones or obstacles, the foot placement regions are constrained and step adaptation is necessary. As step adaptation requires frequent switching between multiple trajectories, an adequate analysis and trajectory switching procedure must be developed. This section develops a trajectory selection policy to run over random stepping stones and jump over obstacles. We employ the trajectory and deadbeat control gain libraries and select a trajectory that satisfies convergence, feasibility, and step location conditions.
\begin{figure}[t!]
\centerline{\includegraphics[width=\columnwidth]{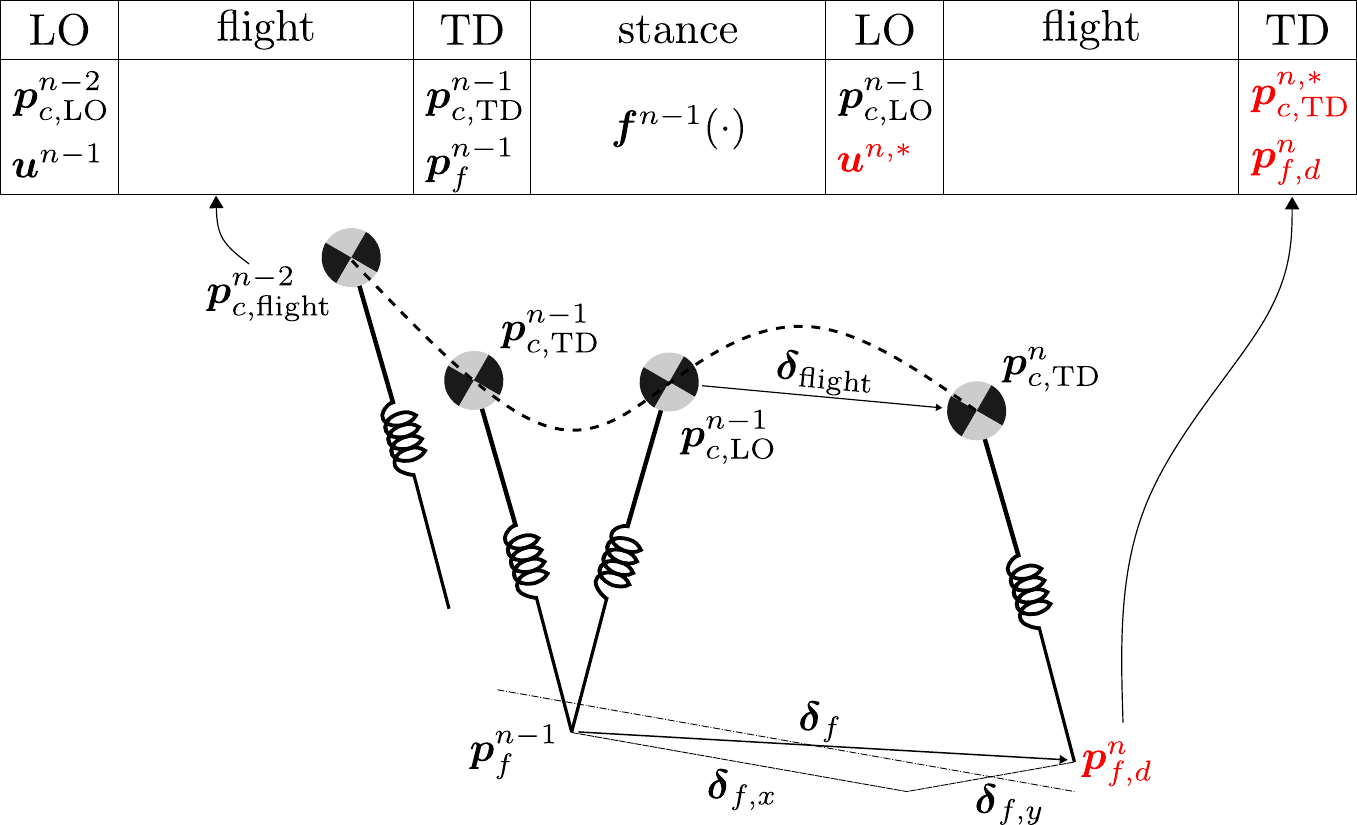}}
\caption{An illustration for multiple-step running with the spring-mass model. The desired stepping configuration is shown in red.}
\label{slipStepFig}
\end{figure}
\subsection{Required Minimum Number of Steps to Plan} \label{stepNumberSec}
The exponential stability \eqref{optimalCLEq} at point-mass dynamics level is guaranteed only if the desired control action $\bm{f}_{d}$ \eqref{fdEq} remains within the feasible set $\mathcal{F}^{3}$. When the feasibility projection in \eqref{QPeqSLIP} is observed, it can be seen that the polyhedral limits, hence the feasibility conditions, are determined by the relative foot placement point with respect to the CoM location. The deadbeat control action in \eqref{slipDeadbeatEq} modifies the periodic foot placement point considering the state errors such that the desired control action $\bm{f}_{d}$ \eqref{fdEq} remains close enough to the feasible set $\mathcal{F}^{3}$ allowing convergence during the stance phase.

Recall the hybrid dynamics for the point-mass model,
\begin{equation}
    \Sigma : 
    \begin{cases}
    \enspace \ddot{\bm{p}}_{c} = \bm{g} + \bm{f}/m, \enspace & ||\bm{r}|| < r_{0} \\
    \enspace \ddot{\bm{p}}_{c} = \bm{g}, \enspace & ||\bm{r}|| \geq r_{0}
    \end{cases},
\end{equation}
and let $(\cdot)_\text{\{TD,LO\}}$ represent the states at the moment of touchdown and liftoff and $\bm{p}_{f,d}^{n}$ be the desired foot placement point for the $n^\text{th}$ step (see Fig.~\ref{slipStepFig}). As stability is closely related to the CoM position and velocity with respect to the foot placement point, foot placement on $\bm{p}_{f,d}^{n}$ must be done with proper center of mass states to preserve the stability and continue to run. Consequently, we target stepping on $\bm{p}_{f,d}^{n}$ with any periodic trajectory $(\bm{x}^{*}, \text{ }\bm{u}^{*})$. To determine the minimum number of steps required for guaranteed convergence, we perform a controllability analysis of the point-mass dynamics and deadbeat control actions through back-integration (see Fig.~\ref{slipStepFig}). 

Assume foot placement on $\bm{p}_{f,d}^{n}$ at the $n^\text{th}$ step with any periodic $(\bm{x}^{*}, \text{ }\bm{u}^{*})$ configuration. The center of mass position at the touchdown is an inverse kinematics function of leg states \eqref{slipFootPointEq},
\begin{equation} \label{step0}
    \bm{p}_{c,\text{TD}}^{n,*} = \text{invkin}(\bm{p}_{f,d}^{n}, \text{ }\bm{u}^{n,*}),
\end{equation}
where $\bm{p}^{*}$ represent configuration states being on the periodic trajectory. The liftoff states of $(n-1)^\text{th}$ are related to the touchdown states of $n^\text{th}$ through the back-integration of ballistic dynamics. Due to the passive dynamics, the states remain on the periodic manifold:
\begin{equation} \label{step1}
    (\bm{p}_{c,\text{LO}}^{n-1,*}, \text{ }\dot{\bm{p}}_{c,\text{LO}}^{n-1,*})
     = \underset{\left[t = \text{TD}^{n} \rightarrow \text{LO}^{n-1}\right]}{\text{integration}}(\Sigma_\text{flight}, \text{ }\bm{p}_{c,\text{TD}}^{n,*}, \text{ }\dot{\bm{p}}_{c,\text{TD}}^{n,*}).
\end{equation}
The touchdown configuration of step $(n-1)$ can be obtained by back-integration through the stance dynamics:
\begin{multline}\label{step2}
    (\bm{p}_{c,\text{TD}}^{n-1}, \text{ }\dot{\bm{p}}_{c,\text{TD}}^{n-1})
     = \underset{\left[t = \text{LO}^{n-1} \rightarrow \text{TD}^{n-1}\right]}{\text{integration}}(\Sigma_\text{stance}, \text{ }\bm{p}_{c,\text{LO}}^{n-1,*}, \text{ }\dot{\bm{p}}_{c,\text{LO}}^{n-1,*},\\ \bm{f}(\bm{f}_{d}^{n-1}(\cdot), \text{ }\bm{p}_{f}^{n-1},\text{ }\mu)).
\end{multline}
The stance phase of step $(n-1)$ in \eqref{step2} is the first phase with the active control force. Consequently, the touchdown states $(\bm{p}_{c,\text{TD}}^{n-1}, \text{ }\dot{\bm{p}}_{c,\text{TD}}^{n-1})$ are not necessarily periodic and convergence to the desired periodic trajectory must be completed during this phase. However, $\bm{p}_{f}^{n-1}$ is already predetermined by the landing configuration of the previous flight phase. As $\bm{p}_{f}^{n-1}$ is related to the feasibility constraints imposed on the control force \eqref{QPeqSLIP}, convergence without deadbeat control \eqref{slipDeadbeatEq} is possible if $\bm{f}_{d}^{n-1}(\cdot)$ remain close enough to the feasible subset $\mathcal{F}^{3}$, but not guaranteed. Consequently, starting to plan from $\text{LO}^{(n-2)}$ and initiating step $\text{TD}^{(n-1)}$ with the deadbeat controlled leg states,
\begin{multline} \label{step3}
   \bm{p}_{f,\text{controlled}}^{n-1}
     = \text{fwdkin}(\underset{\left[t = \text{LO}^{n-2} \rightarrow \text{TD}^{n-1}\right]}{\text{integration}}(\Sigma_\text{flight}, \text{ }\bm{p}_{c,\text{LO}}^{n-2}, \text{ }\dot{\bm{p}}_{c,\text{LO}}^{n-2}),\\ \bm{u}^{*} + \bm{K}\bm{\Delta x})
\end{multline}
results in guaranteed convergence to any periodic trajectory that would land on $\bm{p}_{f,d}^{n}$ on the second step.
\begin{figure}[b!]
\centerline{\includegraphics[width=0.70\columnwidth]{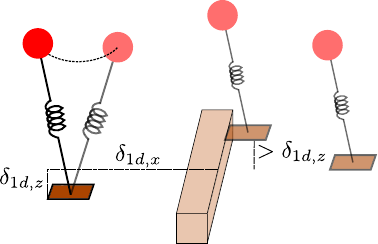}}
\caption{The CoM model jumping over randomly generated obstacles.}
\label{randomObstaclesFig}
\end{figure}
\subsection{Jumping Over Randomly Appearing Obstacles}
Let some random obstacles appear in front of the robot immediately after touchdown, so that the robot must jump over them and continue running, maintaining its stability. Let $\delta_{1d,z}$ be the height of the obstacle and the end of the obstacle is $\delta_{1d,x}$ away from the back of the foot (see Fig.~\ref{randomObstaclesFig}). As the obstacle appears immediately after touchdown, planning begins from \eqref{step2}, targeting stepping on any $\bm{p}_{f}^{n}$ that lies behind the object. As $\bm{p}_{f}^{n-1}$ is predetermined by the previous flight phase, convergence to any trajectory is not guaranteed but still possible for some, depending on the overlap between the desired forces $\bm{f}_{d,i}(\cdot)$ and the feasibility set $\mathcal{F}^{3}_{i}$ for $i^\text{th}$ trajectory. Let $\bm{x}_\text{TD}$ and $\bm{u}_\text{TD}$ be the actual spring-mass states of the system at the touchdown moment. Recalling from \eqref{slipDeadbeatEq}, if the touchdown leg states are close enough to the deadbeat controlled leg states for $i^\text{th}$ trajectory,
\begin{equation} \label{jumpConv}
    \bm{u}_\text{TD} \approx \bm{u}^{*}_{i} + \bm{K}_{i} (\bm{x}_\text{TD} - \bm{x}^{*}_{i}),
\end{equation}
convergence is possible in the given stance phase. The trajectory selection policy employs \eqref{jumpConv} and comprises two complementary parts: touchdown (see Alg.~\ref{alg1}) and liftoff (see Alg.~\ref{alg2}). Both algorithms output the same trajectory when no disturbance is present. On the other hand, repetitive re-selection is done for re-planning purposes to address disturbances and uncertainties. The touchdown algorithm, Alg.~\ref{alg1}, first gathers all trajectories that satisfy the jumping conditions and picks the one with the highest convergence possibility with respect to \eqref{jumpConv}. The liftoff algorithm, Alg.~\ref{alg2}, updates the selection in the case of disturbance, uncertainty, or incomplete convergence conditions. The algorithm seeks a trajectory that satisfies the landing conditions and maintains stability after landing. As multiple trajectories may satisfy conditions in Alg.~\ref{alg2}, one with the closest apex states is chosen for minimal convergence effort.
\begin{algorithm} [t!]
\small
\caption{Jumping over obstacles: the touchdown policy.}\label{alg1}
\begin{enumerate} [wide, labelwidth=!,itemindent=!,labelindent=0pt, leftmargin=0em]
    \item Eliminate all trajectories whose leg clearance at the apex height is less than $\delta_{1d,z}$. Recalling \eqref{slipFootPointEq}, the elimination condition follows:
    \begin{equation*}
        \delta_{1d,z} < p_{c,z,\mathrm{apex},i} - l_{h}\cos{\theta_{1,i}}\cos{\theta_{2,i}}.
    \end{equation*}
    \item Calculate foot displacements using \eqref{stepLength} for the case of full convergence and store them in a library $\bm{\Delta} = \begin{bmatrix} \bm{\delta}_{1}^{\top} & \bm{\delta}_{2}^{\top} & \dots \end{bmatrix}^{\top}$.
    \item Reduce the displacement library $\bm{\Delta}$ with
    \begin{equation*}
        \delta_{1d,x} < \bm{\Delta}_{x}.
    \end{equation*}
    \item Pick a trajectory with the highest possibility of convergence:
    \begin{equation*}
        \min_{i}||\bm{u}_\text{TD} - (\bm{u}^{*}_{i} + \bm{K}_{i} (\bm{x}_\text{TD} - \bm{x}^{*}_{i}))||.
    \end{equation*}
\end{enumerate}
\end{algorithm}
\begin{algorithm} [t!]
\small
\caption{Jumping over obstacles: the liftoff policy.}\label{alg2}
\begin{enumerate} [wide, labelwidth=!,itemindent=!,labelindent=0pt, leftmargin=0em]
    \item Estimate the apex state $\bm{x}_{\text{apex}}$ of the free-flying ballistic dynamics. For a rise time $t_{\text{rise}} = v_{z}/g$ definition, the apex height is simply: 
    \begin{equation*}
        h_{\text{apex}} = p_{\text{com},z} + v_{z}t_{\text{rise}} - 0.5gt_{\text{rise}}^2.
    \end{equation*}
    \item Recall \eqref{slipDeadbeatEq} and calculate the deadbeat-controlled new set of leg states for guaranteed convergence to any trajectory:
    \begin{equation*}
        \bm{u}_{i,\text{new}} = \bm{u}^{*}_{i} + \bm{K}_{i} (\bm{x}_{\text{apex}} - \bm{x}^{*}_{i}).
    \end{equation*}
    \item Eliminate all kinematically infeasible trajectories:
    \begin{equation*}
        \begin{gathered}
            l_{h,min} < l_{h,\mathrm{new},i} < l_{h,max}, \\
            \theta_{1,min} < \theta_{1,\mathrm{new},i} < \theta_{1,max}, \\
            \theta_{2,min} < \theta_{2,\mathrm{new},i} < \theta_{2,max}. \\
        \end{gathered}
    \end{equation*}
    \item Eliminate all trajectories whose leg clearance at the apex height is less than $\delta_{1d,z}$. Recalling \eqref{slipFootPointEq}, the elimination condition follows:
    \begin{equation*}
        \delta_{1d,z} < p_{c,z,\mathrm{apex},i} - l_{h,\mathrm{new},i}\cos{\theta_{1,\mathrm{new},i}}\cos{\theta_{2,\mathrm{new},i}}.
    \end{equation*}
    \item Calculate the possible foot placement points \eqref{stepLength} with the new leg states $\bm{u}_{i,\text{new}}$ and store them: $\bm{\Delta} = \begin{bmatrix} \bm{\delta}_{1}^{\top} & \bm{\delta}_{2}^{\top} & \dots \end{bmatrix}^{\top}$.
    \item Reduce the displacement library $\bm{\Delta}$ with
    \begin{equation*}
        \delta_{1d,x} < \bm{\Delta}_{x}.
    \end{equation*}
    \item Pick a trajectory with minimal convergence effort:
    \begin{equation*}
        \min_{i}||\bm{x}_\text{apex} - \bm{x}^{*}_{i}||.
    \end{equation*}
\end{enumerate}
\end{algorithm}
\subsection{Running Through Randomly Generated Stepping Stones}
\begin{figure}[t!]
\centerline{\includegraphics[width=0.90\columnwidth]{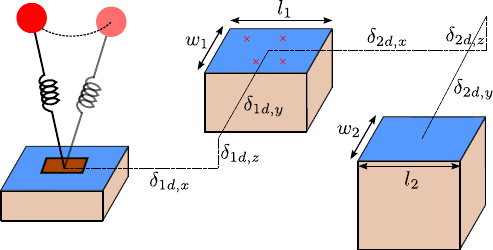}}
\caption{The CoM model navigating over stepping stones. The red crosses represent possible random foot placement points due to the relaxation conditions, which require an update on $\bm{\delta}_{2d}$ for the next step.}
\label{steppingStonesFig}
\end{figure}

Let the following two stepping stones are separated by $\bm{\delta}_{1d} \in \mathbb{R}^{3}$ and $\bm{\delta}_{2d} \in \mathbb{R}^{3}$ (see Fig.~\ref{steppingStonesFig}). Furthermore, let $\bm{\delta}_{r} = \begin{bmatrix} l/2 & w/2 \end{bmatrix}^{\top}$ be a relaxation such that the foot does not have to be on the center but is still inside the stepping region. Due to strict foothold constraints, planning starts from \eqref{step3}, corresponding to step $(n-2)$, so that landing step $(n-1)$ with the deadbeat controlled leg states results in guaranteed convergence to the desired trajectory to land on $\bm{p}_{f,d}^{n}$. Steps $(n-2)$, $(n-1)$, and $(n)$ represent the current foot location, one-ahead stepping stone, and two-ahead stepping stone, respectively.

The selection policy is triggered at each touchdown and liftoff, resulting in repetitive re-selections to address disturbances and uncertainties. The touchdown and liftoff selection policies are shown in Alg.~\ref{alg3} and Alg.~\ref{alg4} and are extensions of Alg.~\ref{alg1} and Alg.~\ref{alg2}, respectively. The touchdown selection policy targets stepping on the next stepping stone. The liftoff selection policy targets stepping anywhere on the one-ahead stepping stone with the deadbeat controlled leg states, such that converging onto the suitable trajectory results in stepping on the two-ahead stepping stone with a periodic trajectory.
\begin{figure}[b!]
\centerline{\includegraphics[width=0.70\columnwidth]{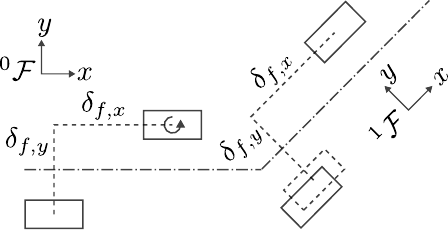}}
\caption{Foot placement illustration when the robot changes the running direction suddenly at the second step. During the stance phase, the system aims to converge to the desired trajectory. During the flight phase, the deadbeat controller reacts to the remaining errors by shifting the foot placement point towards the bottom right of the nominal.}
\label{directionChangeFig}
\end{figure}
\begin{algorithm}[t!]
\small
\caption{Random stepping stones: the touchdown policy.}\label{alg3}
\begin{enumerate} [wide, labelwidth=!,itemindent=!,labelindent=0pt, leftmargin=0em]
    \item  Apply Alg.~\ref{alg1} for the next step with the following two modifications:
    \begin{itemize}
        \item In step 2, take account of surface-level differences and calculate the flight time ($t_{\text{flight}} = t_{\text{rise}} + t_{\text{fall}}$) accordingly.
        \item In step 3, the reduction condition changes:
            \begin{equation*}
            \begin{gathered}
                \delta_{1d,x} - w_{1}/2 < \bm{\Delta}_{x} < \delta_{1d,x} + w_{1}/2, \\
                \delta_{1d,y} - l_{1}/2 < \bm{\Delta}_{y} < \delta_{1d,y} + l_{1}/2.
            \end{gathered}
            \end{equation*}
    \end{itemize}
\end{enumerate}
\end{algorithm}
\begin{algorithm} [t!]
\small
\caption{Random stepping stones: the liftoff policy.}\label{alg4}
\begin{enumerate} [wide, labelwidth=!,itemindent=!,labelindent=0pt, leftmargin=0em]
    \item Apply Algorithm~\ref{alg2} to find all trajectories that end up stepping on the one-ahead stepping stone with the following modifications:
    \begin{itemize}
        \item In step 5, take account of surface-level differences and calculate the flight time ($t_{\text{flight}} = t_{\text{rise}} + t_{\text{fall}}$) accordingly.
        \item In step 6, the reduction condition changes:
            \begin{equation*}
            \begin{gathered}
                \delta_{1d,x} - w_{1}/2 < \bm{\Delta}_{x} < \delta_{1d,x} + w_{1}/2, \\
                \delta_{1d,y} - l_{1}/2 < \bm{\Delta}_{y} < \delta_{1d,y} + l_{1}/2.
            \end{gathered}
            \end{equation*}
        \item Skip step 7.
    \end{itemize}
\end{enumerate}
\begin{enumerate} [wide, labelwidth=!,itemindent=!,labelindent=0pt, leftmargin=0em] \setcounter{enumi}{1}
    \item Eliminate all trajectories whose leg clearance at the apex height is less than $\delta_{2d,z}$.
    \item As a result of the relaxation conditions, each trajectory has a different foot placement point on the first stepping stone (see Fig.~\ref{steppingStonesFig}). Update $\delta_{2d,x}$ and $\delta_{2d,y}$ for each remaining trajectory:
    \begin{equation*}
    \begin{gathered}
        \delta_{2d,x,i} = \delta_{2d,x} - (\delta_{i,x} - \delta_{2d,x}), \\
        \delta_{2d,y,i} = \delta_{2d,y} + (\delta_{i,y} - \delta_{2d,y}).
    \end{gathered}
    \end{equation*}
    \item Calculate foot displacements using \eqref{stepLength} for the case of full convergence, also considering the height differences and flight times accordingly, and reconstruct the displacement library $\bm{\Delta}$.
    \item Further reduce the displacement library $\bm{\Delta}$ for the following two relaxed conditions:
    \begin{equation*}
    \begin{gathered}
        \delta_{2d,x,i} - w_{2}/2 < \bm{\Delta}_{x,i} < \delta_{2d,x,i} + w_{2}/2, \\
        \delta_{2d,y,i} - l_{2}/2 < \bm{\Delta}_{y,i} < \delta_{2d,y,i} + l_{2}/2.
    \end{gathered}
    \end{equation*}
    \item Pick a trajectory with minimal convergence effort:
    \begin{equation*}
        \min_{i}||\bm{x}_\text{apex} - \bm{x}^{*}_{i}||.
    \end{equation*}
\end{enumerate}
\end{algorithm}
\subsection{Changing the Running Direction}
Changing the heading angle combines two tasks: rotating the inertia/body and changing the CoM velocity. This section assumes body rotations are handled at the whole-body control level and focuses on CoM control. During running, rotating the body while failing to change the CoM velocity direction causes the robot to roll and fall to the sides (Fig.~\ref{firstPageFig}C shows the body lean and the tendency to roll side when a sudden turn is applied while running). Therefore, step adaptation is required for the next step through the deadbeat controller.

Define $\bm{R}_{h} \in \mathbb{R}^{3 \times 3}$ to be a rotation matrix from the world frame $^{0}\mathcal{F}$ to the new frame $^{1}\mathcal{F}$ aligned with the new desired running direction (see Fig.~\ref{directionChangeFig}). The reference CoM trajectories, $\bm{p}_{c,d}$, $\dot{\bm{p}}_{c,d}$, and $\ddot{\bm{p}}_{c,d}$ are then transformed into the new frame, i.e., $^{1}\bm{p}_{c,d} = \bm{R}_{h}{^{0}\bm{p}_{c,d}}$. Similarly, the error definition for the deadbeat controller \eqref{slipDeadbeatEq} is transformed back to the world frame such that $\bm{\Delta x} = \bm{R}_{h}^{\top}{^{1}\bm{x}} - \bm{x}^{*}$.
\subsection{Scalability of the Policy}
The trajectory selection algorithms involve combinations of scalar and 3D vector-matrix algebra operations for each trajectory. Hence, the computation time is linearly proportional to the number of trajectories in the library. We measure $20\mu s$ to compute the most expensive algorithm, Alg.~\ref{alg4}, for the example library used in this study, which contains 315 trajectories (see Fig.~\ref{libraryFig}). Consequently, the number of trajectories can reach more than ten thousand satisfying the real-time constraint.
\section{Implementation on Humanoid Robots}
Humanoid robots are high-degree-of-freedom complex floating base systems. They include multiple subsystems, such as legs, arms, head, and other joints on the torso and pelvis. Control of humanoid robots includes a set of task definitions covering the whole-body dynamics, such as swinging the legs and arms, keeping the torso upright, and following a CoM trajectory. The task formulations can be based on different controllers, such as inverse dynamics control as in this paper, or passivity-based control \cite{passivity-wbc0, passivity-wbc1, mptc}. Then, the tasks are combined in a QP problem and solved all together to find a solution that satisfies all the tasks the best \cite{wbc0, wbc1, wbc2}. Task prioritization can be handled either strictly by projecting lower-priority tasks into the null space of high-priority tasks \cite{wbc0,wbc1,passivity-prioritization} or softly with simple weight adjustments, as in this paper. Detailed WBC formulations with strict and soft prioritization are provided in \cite{wensingPhD} and \cite{JohannesPhD}, respectively.

This section combines the conventional inverse dynamics whole-body control formulation with two add-ons for collision avoidance and reactive limb swing control. The new task definitions introduce adaptiveness to the WBC formulation, enabling it to handle multiple CoM trajectories without requiring additional tuning of gains and limb-swing trajectories. The reactive limb swing control is relevant and important in the case of frequent trajectory switching at the CoM level, resulting in a wide variation of stance and flight phase characteristics. The collision detection and avoidance prevent reactive limb swinging from resulting in self-collision and can be implemented to prevent possible arm-to-body and leg-to-leg collisions.
\subsection{Floating Base System Dynamics}
Let $\bm{q}$ be a set of configuration variables and $\bm{\nu} = (\bm{\nu}_{b}, \bm{\nu}_{j})$ be the generalized velocity where $\bm{\nu}_{b} = (\bm{v}_{b},\bm{\omega}_{b}) \in \mathbb{R}^{6}$ is the linear and angular velocity of the floating base and $\bm{\nu}_{j} \in \mathbb{R}^{n}$ is the generalized velocity of the joints. The well-known robotic system dynamics results in
\begin{equation} \label{generalDyn}
\bm{M}(\bm{q}) \dot{\bm{\nu}}
+ \bm{C}(\bm{q},\bm{\nu})\bm{\nu} + \bm{\tau}_{g}(\bm{q}) = 
\begin{bmatrix}
    \bm{0}\\
    \bm{\tau}
\end{bmatrix}
+
\bm{J}_{c}(\bm{q})^{\top} \bm{f}_{c}+
\bm{J}_{p}(\bm{q})^{\top} \bm{f}_{p},
\end{equation}
where $\bm{M}$ is inertia matrix; $\bm{C}$ is Coriolis matrix; $\bm{\tau}_{g}$ is gravity vector; $\bm{\tau}$ is joint torques; $\bm{J}_{c}$ is contact Jacobians; $\bm{f}_{c}$ is contact forces; $\bm{J}_{p}$ is closed chain constraint Jacobians; $\bm{f}_{p}$ is closed chain constraint forces. In the case of closed kinematic linkages, the passive joints are constrained to move together with their corresponding actively driven counterparts through holonomic constraints \cite{sovukluk_wbc, sovukluk_wbc2}.
\subsection{Task Formulation via Inverse Dynamics Control (IDC)}
For a given task Jacobian $\bm{J}_{i}$, the task space velocity yields
\begin{equation} \label{taskSpaceVel}
    \dot{\bm{x}}_{i} = \bm{J}_{i}\bm{\nu}.
\end{equation}
The time derivative of \eqref{taskSpaceVel} yields the task acceleration,
\begin{equation} \label{taskSpaceAcc}
    \ddot{\bm{x}}_{i} = \bm{J}_{i}\dot{\bm{\nu}} + \dot{\bm{J}}_{i}\bm{\nu}.
\end{equation}
Setting $\ddot{\bm{x}}_{i} = \ddot{\bm{x}}_{i,d}$, where
\begin{equation} \label{desAccID}
    \ddot{\bm{x}}_{i,d} = \ddot{\bm{x}}_{i,ref} + \bm{K}_{D,i}(\underbrace{\dot{\bm{x}}_{i,ref} - \dot{\bm{x}}_{i}}_{\dot{\bm{e}}}) + \bm{K}_{P,i}(\underbrace{\bm{x}_{i,ref} - \bm{x}_{i}}_{\bm{e}_{i}})
\end{equation}
and $\{\bm{K}_{D,i}, \bm{K}_{P,i}\}$ are positive definite gain matrices, the closed-loop error dynamics results in being exponentially stable,
\begin{equation} \label{CLID}
    \ddot{\bm{e}}_{i} + \bm{K}_{D,i}\dot{\bm{e}}_{i} + \bm{K}_{P,i}\bm{e}_{i} = \bm{0}.
\end{equation}
\begin{figure}[b!]
\centerline{\includegraphics[width=0.50\columnwidth]{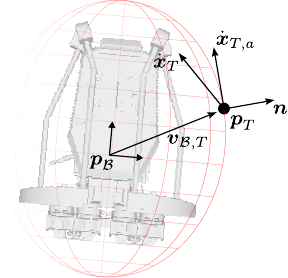}}
\caption{An illustration of a task control reaching the ellipsoidal boundary. The task velocity $\dot{\bm{x}}_{T}$ is projected on the tangent space to prevent collision.}
\label{collisionFig}
\end{figure}
\subsection{Collision Detection and Avoidance}
Let $E: \mathbb{R}^{3}\rightarrow\mathbb{R}$ be a general ellipsoid function,
\begin{equation}
    E(\bm{p}) = \dfrac{p_{x}}{a^{2}} + \dfrac{p_{y}}{b^{2}} + \dfrac{p_{z}}{c^{2}} - 1,
\end{equation}
where $\{a,b,c\} \in \mathbb{R}$ are radius of each axes. Consequently, 
\begin{equation*}
    \mathcal{E} = \{ \bm{p}\in\mathbb{R}^{3} \enspace | \enspace E(\bm{p}) = 0 \}
\end{equation*}
collects all the points on the ellipsoidal surface. Assume a task $T$ is placed at point $\bm{p}_{T}$ and a vector $\bm{v}_{\mathcal{B},T} = \bm{p}_{T} - \bm{p}_{\mathcal{B}}$ defines a vector from the ellipsoidal's center to the task. Let $\bm{n}$ be the surface normal. In the case $E(\bm{v}_{\mathcal{B},T}) \leq 0$ and $\dot{\bm{x}}_{T}^{\top} \bm{n} < 0$, implying, the task is inside the ellipsoid and its velocity towards inside, respectively, the task velocity must be projected to prevent collision (see Fig.~\ref{collisionFig}). Projection onto the nullspace of the ellipsoid's surface normal results in movement in the tangential direction. The allowed velocity $\dot{\bm{x}}_{T,a}$ is simply
\begin{equation} \label{projection}
    \dot{\bm{x}}_{T,a} = \dot{\bm{x}}_{T} - \mathrm{proj}_{\bm{n}}(\dot{\bm{x}}_{T}),
\end{equation}
where
\begin{equation*}
    \bm{n} = \dfrac{\mathrm{d}E}{\mathrm{d}\bm{q}} \bigg\rvert _{\bm{p} = \bm{p}_{T}}.
\end{equation*}
The resultant task velocity is introduced as a high-priority task in WBC formulation. In case $E(\bm{v}_{\mathcal{B},T}) > 0$ or $E(\bm{v}_{\mathcal{B},T}) \leq 0$ but $\dot{\bm{x}}_{T}^{\top} \bm{n} > 0$ no action is necessary as the task is either outside of the region, or inside of the region but moving towards outside.
\subsection{Reactive Limb Swing Control}
Another important aspect of humanoid robot locomotion is torso and limb swing control during running. As the 3D spring-mass dynamics covers only the CoM control and the foot placement point, the remaining limb swing and posture control must be handled at the WBC level. Independent of the CoM stability, preserving the postural stability, e.g., keeping the body upright, is crucial for the continuity of the motion. As this study combines varying running behaviors, in terms of trajectories, stance time, and flight time, the limb-swing and torso control must be generic and suitable for all.
\subsubsection{Orientational Dynamics and Dynamic Coupling}
The centroidal momentum is related to the generalized velocities through 
\begin{equation}
    \bm{h}_{G} = \bm{A}_{G}(\bm{q}) \bm{\nu},
\end{equation}
where $\bm{A}_{G} \in \mathbb{R}^{6 \times (n_j+6)}$ is the centroidal momentum matrix (CMM) \cite{CMM1,CMM2} and $n_j$ is the number of joints. The CMM is composed of linear and angular parts,
\begin{equation}
    \bm{A}_{G} = 
    \begin{bmatrix}
        \bm{A}_{l} \in \mathbb{R}^{3 \times (n_j+6)} \\
        \bm{A}_{k} \in \mathbb{R}^{3 \times (n_j+6)}
    \end{bmatrix}.
\end{equation}
Furthermore, the centroidal angular momentum matrix (CAMM) $\bm{A}_{k}$ can also be decomposed into
\begin{equation} \label{Ak}
    \bm{A}_{k} = 
    \begin{bmatrix}
        \bm{A}_{v}\in\mathbb{R}^{3 \times 3} & \bm{A}_{\omega}\in\mathbb{R}^{3 \times 3} & \bm{A}_{j}\in\mathbb{R}^{3 \times n_j}
    \end{bmatrix},
\end{equation}
where $\bm{A}_{v}$, $\bm{A}_{\omega}$, and $\bm{A}_{j}$ represent body translational, body rotational, and joint velocity portions, respectively. Among the submatrices of the CAMM, $\bm{A}_{\omega}$ is invertible and $\bm{A}_{v}$ is always a zero matrix for all configurations $\forall \bm{q}$ \cite{sovukluk2025_icra}. The centroidal angular momentum is then equal to
\begin{equation} \label{CAM}
    \bm{k}_{G} = \begin{bmatrix} \bm{A}_{\omega} & \bm{A}_{j} \end{bmatrix}
    \begin{bmatrix} \bm{\omega}_{b} \\ \bm{\nu}_{j} \end{bmatrix}.
\end{equation}
Reorganizing \eqref{CAM} results in the base frame orientation dynamics showing the coupling between the joint velocities and base frame rotational velocity,
\begin{equation} \label{baseFrameDyn}
    \bm{\omega}_{b} = \bm{A}_{\omega}^{-1}(\bm{k}_{G}-\bm{A}_{j}\bm{\nu}_{j}).
\end{equation}
Furthermore, the rate of centroidal angular momentum,
\begin{equation} \label{CMrate}
        \dot{\bm{k}}_{G}
    = 
    \sum_{i=1}^{n_{c}}
        \left((\bm{p}_{\text{contact},i} - \bm{p}_{\text{CoM}}) \times \bm{f}_{\text{contact},i}\right),
\end{equation}
is related to the interaction forces at the contact points \cite{holonomy, CMM2}. During the flight phase, where the contact forces are zero, the centroidal angular momentum is conserved, $\dot{\bm{k}}_{G} = \bm{0}$. Consequently, the base orientation cannot be controlled explicitly, but is a function of the constant angular momentum and joint velocities \eqref{baseFrameDyn}. During the stance phase, however, the problem of orientation control is easier as it can be controlled explicitly by the ground reaction forces \eqref{CMrate}.
\subsubsection{Control Action}
The stance phase includes multiple tasks: tracking the given CoM trajectory originating from the spring-mass model, keeping the stance foot fixed on the ground, keeping the torso upright, swinging the other foot forward to prepare for the next step, and swinging the arms. Among the stance phase tasks, arm swing is a redundant task that can be utilized in a reactive manner to support postural stability. The flight phase, on the other hand, is more redundant and harder to control than the stance phase, as no foot is fixed on the ground. The only control target originating from the spring-mass model is placing the upcoming stance foot at a certain location with respect to the CoM and waiting for touchdown. The remaining joints, which are the swing leg and arms, must be controlled for the postural stability to prevent any excessive roll due to the conservation of angular momentum \eqref{CMrate} and coupled body velocity dynamics \eqref{baseFrameDyn}.

Let $n_{e}$ and $n_{r}$ be the number of essential and redundant joints such that $n_{j} = n_{e} + n_{r}$. Furthermore, let
\begin{equation}
    \begin{bmatrix}
        \bm{J}_{v} \\
        \bm{J}_{\omega} \\
        \bm{J}_{e} \\
        \bm{J}_{r}
    \end{bmatrix}
    =
    \begin{bmatrix}
        \bm{I}_{3 \times 3} & \bm{0}_{3 \times 3} & \bm{0}_{3 \times n_{e}} & \bm{0}_{3 \times n_{r}} \\
        \bm{0}_{3 \times 3} & \bm{I}_{3 \times 3} & \bm{0}_{3 \times n_{e}} & \bm{0}_{3 \times n_{r}} \\
        \bm{0}_{n_{e} \times 3} & \bm{0}_{n_{e} \times 3} & \bm{I}_{n_{e} \times n_{e}} & \bm{0}_{n_{e} \times n_{r}} \\
        \bm{0}_{n_{r} \times 3} & \bm{0}_{n_{r} \times 3} & \bm{0}_{n_{r} \times n_{e}} & \bm{I}_{n_{r} \times n_{r}} \\
    \end{bmatrix}
\end{equation}
collect all joint-level Jacobians for body translational, body rotational, essential joints, and redundant joints, respectively. The design objective is controlling the base frame orientation task related to $\bm{J}_{\omega}$, in the nullspace of $\bm{J}_{t} = [\bm{J}_{v};\bm{J}_{e}]$, such that only redundant joints are utilized. Let
\begin{equation}
    \bm{N}_{t} = \bm{I} - \bar{\bm{J}}_{t} \bm{J}_{t}
\end{equation}
be the nullspace projection of $\bm{J}_{t}$ where $\bar{\bm{J}}_{t}$ is dynamically consistent pseudo-inverse of $\bm{J}_{t}$ \cite{wbc2}. The new task Jacobian for base orientation control becomes
\begin{equation} \label{bodyProjectedJac}
    \bm{J}_{\omega,p} = \bm{J}_{\omega}\bm{N}_{t}.
\end{equation}
Defining a WBC task \eqref{taskSpaceAcc} through $\bm{J}_{\omega,p}$ and minimizing for the desired acceleration \eqref{desAccID} written in terms of actual base orientational position and velocity errors results in a control through the ground reaction forces (if available) and redundant joints.

Note that the proposed task formulation for base orientation control is not a task prioritization but rather a joint isolation, as they are written at the joint level. The formulation allows only a certain set of joints to be utilized for body orientation control, regardless of other tasks. In case multiple task formulations share common joints, they are still affected by each other and prioritized by weight matrices. A task written with $\bm{J}_{\omega,p}$ for the base orientation control results in inherently determined redundant joint control for the stance and flight phases.
\begin{figure*}[t]
\centerline{\includegraphics[width=\textwidth]{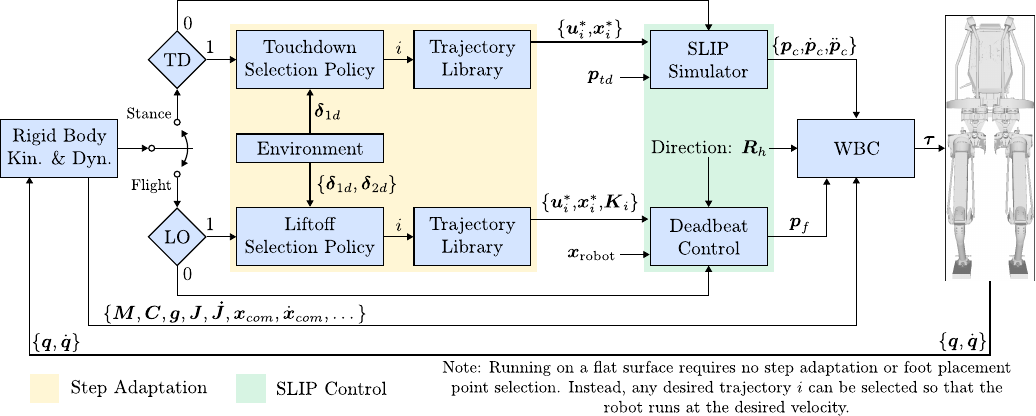}}
\caption{Overall control system diagram for the proposed method. Note that the activated parts differ for the stance and flight phases.}
\label{systemDiagramFig}
\end{figure*}
\subsection{Whole Body Control Formulation}
The desired CoM and limb trajectories are mapped to a humanoid robot through an optimization-based whole-body controller. Selecting the parameters to optimize for as $\{\dot{\bm{\nu}}; \bm{\tau}; \bm{f}_{c}; \bm{f}_{p}\}$, the combination of all task controllers \eqref{desAccID} into a quadratic problem constitutes the cost function of the problem:
\begin{subequations} \label{QPeqID}
    \begin{equation} \tag{\ref{QPeqID}}
    \underset{\ddot{\bm{q}}, \bm{\tau}, \bm{f}_{c}, \bm{f}_{p}}{min} \sum_{i}{(\bm{\ddot{x}}_{i,d} -\bm{\ddot{x}}_{i})^{\top} \bm{W}_{i}(\bm{\ddot{x}}_{i,d} - \bm{\ddot{x}}_{i})}
    \end{equation}
    \begin{equation*}
        \text{Such that:}
    \end{equation*}
    \begin{equation} \label{c1}
    \bm{M} \dot{\bm{\nu}}
+ \bm{C}\bm{\nu} + \bm{\tau}_{g} = 
\begin{bmatrix}
    \bm{0}\\
    \bm{\tau}
\end{bmatrix}
+
\bm{J}_{c}^{\top} \bm{f}_{c}+
\bm{J}_{p}^{\top} \bm{f}_{p},
    \end{equation}
    \begin{equation} \label{c2}
    \bm{J}_{c}\dot{\bm{\nu}} + \dot{\bm{J}}_{c}\dot{\bm{q}} = 0,
    \end{equation}
    \begin{equation} \label{c3}
    \left| f_{x,l} \right| \leq \dfrac{\mu f_{z,l}}{\sqrt{2}}\text{,} \enspace \left| f_{y,l} \right| \leq \dfrac{\mu f_{z,l}}{\sqrt{2}}\text{,} \enspace \text{and} \enspace f_{z} \geq 0 \enspace \forall l,
    \end{equation}
    \begin{equation} \label{c4}
    \bm{J}_{p}\dot{\bm{\nu}} + \dot{\bm{J}}_{p}\dot{\bm{q}} = 0,
    \end{equation}
    \begin{equation} \label{c5}
    \bm{\tau_{min}} \leq \bm{\tau} \leq \bm{\tau_{max}},
    \end{equation}
\end{subequations}
where \eqref{c2} and \eqref{c3} are stationary foot and friction constraints. As the cost function \eqref{QPeqID} is constituted by a combination of different tasks and their corresponding weight matrices $\bm{W}_{i}$, the tasks with higher weights are prioritized.
%
%
\begin{figure*} [t!]
\centerline{\includegraphics[width=\textwidth]{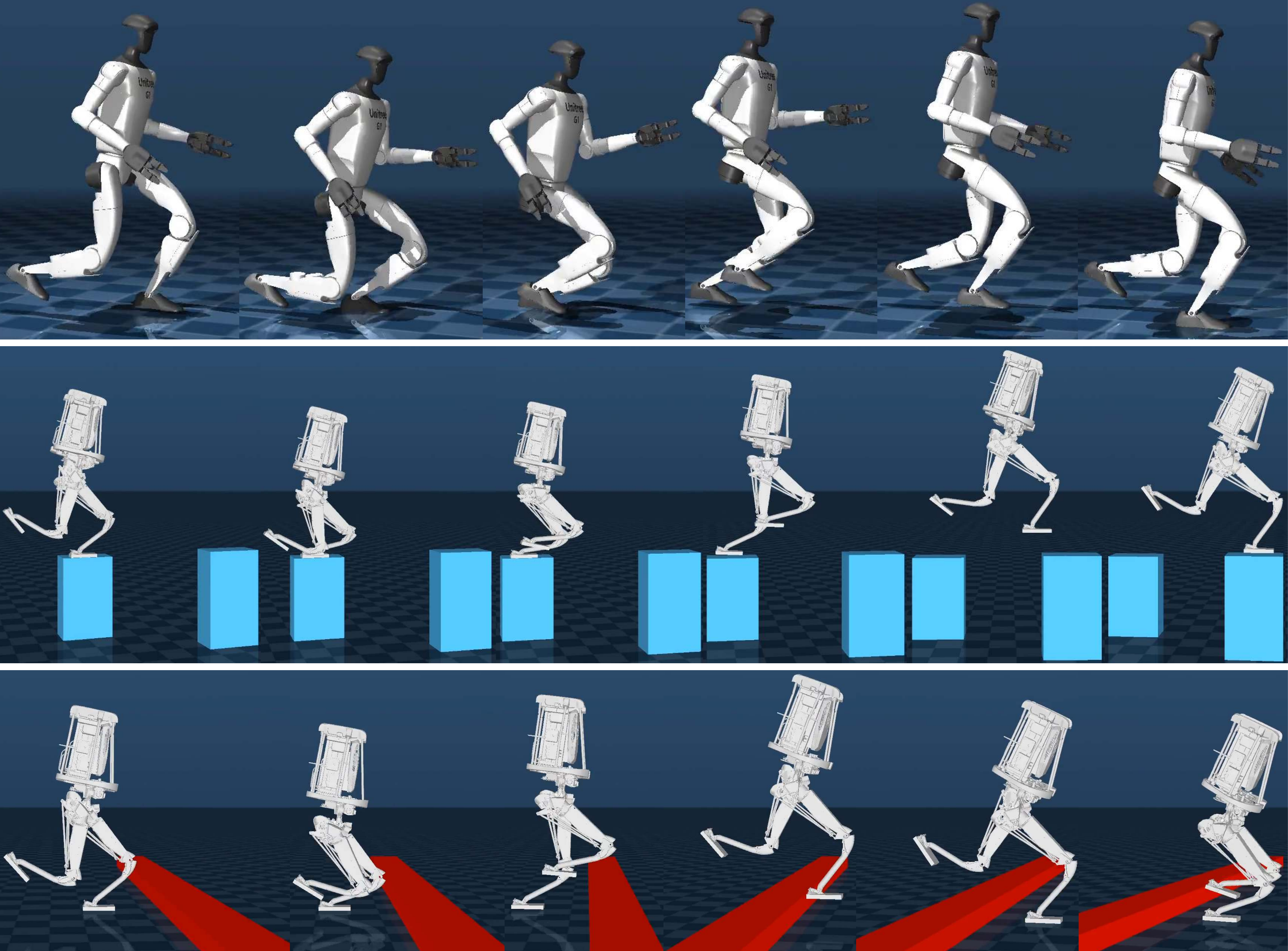}}
\caption{Top: Unitree G1 running at $1 m/s$. Middle: Kangaroo running through randomly generated stepping stones which are 60 to 100, 35 to 45, and $\pm 10$ centimeters apart from each other in the longitudinal, lateral, and vertical directions, respectively. Bottom: Jumping over randomly generated obstacles that appear 40 centimeters ahead. The objects' width and height vary between 5 to 30 and 10 to 15 centimeters, respectively. The remaining behaviors are shown in Fig.~\ref{firstPageFig}.}
\label{flatRunFig}
\end{figure*}
\section{Simulations}
The simulation results include flat surface running with velocity changes, sudden turns, and running through stepping stones. To further justify the proposed method’s robustness against real-world challenges, we also simulate its operation under a comprehensive and significant set of signal noise, imprecision, and delays. The supplemental video includes all and more, including responses to force disturbances, unobserved height differences, edge stepping, and slalom running. Note that, although this paper presents diverse and highly dynamic behaviors (see Figs.~\ref{firstPageFig}~and~\ref{flatRunFig}), all behaviors were obtained using the same parameter set without requiring additional tuning, thanks to the reactive limb-swing formulation.

We use MuJoCo Physics Simulator \cite{mujoco} for simulation; Pinocchio \cite{pinocchio} for rigid body dynamics and kinematics calculations; ProxQP to solve the QP problems in \eqref{QPeqID} and \eqref{QPeqSLIP}; Ceres Solver \cite{ceres} to solve the nonlinear least-squares problems for periodic trajectory optimization \eqref{slipSearchEq}. We set the friction coefficient $\mu = 0.6$, considering concrete terrains. Execution times for control and library calculations are provided in Table~\ref{timeTab}.

In addition to reactive limb swing control, we introduce a low-weight zero position and velocity task for redundant joints to enhance numerical stability. The limb-swing task overrides the low-weight zero configuration task, causing the redundant limbs to swing inherently to maintain the torso's upright position, as shown in Fig.~\ref{flatRunFig}. Kangaroo uses backward leg swing for angular momentum compensation (no arms), while Unitree G1 employs both leg and arm swing — validating the framework's adaptability to kinematic variability. The backward leg swing is more significant for Kangaroo as it does not have any arms to compensate for angular momentum. Nevertheless, even without arms, the control system manages to handle long flight phases as shown in Fig.~\ref{flatRunFig}. The reactive control also significantly enhances disturbance rejection capabilities at the whole-body control level.

\begin{table}[b!] \centering \caption{Execution times where $T$ and $\#$ represent the sampling time and number of repetitions. CPU: AMD Ryzen 7 5800X, Interface: C++.}
\begin{tabular}{ccccc}
Offline & & $T$ & $\#$ & time \\ \hline
       & SLIP Library \eqref{slipSearchEq} & $10^{-4}$ & $315$ & $1.2s$ \\ \cline{2-2}
       & $\bm{J}_{x}$ of active SLIP \eqref{slipTaylorEq} & $10^{-3}$ & 315 & $1.6s$ \\ \cline{2-2}
       & $\bm{J}_{u}$ of active SLIP \eqref{slipTaylorEq} & $10^{-3}$ & 315 & $1.6s$ \\ 
Online &  & \multicolumn{3}{c}{time} \\ \hline
       & \begin{tabular}{c} Trajectory Selection \\ with $315$ Trajectories\end{tabular} & \multicolumn{3}{c}{$20$ $\mu s$}\\ \cline{2-2}
       & WBC Formulation & \multicolumn{3}{c}{$35$ $\mu s$} \\ \cline{2-2}
       & WBC Solution Avg. & \multicolumn{3}{c}{$125$ $\mu s$} 
\end{tabular}
\label{timeTab}
\end{table}
\begin{figure*} [t!]
\centerline{\includegraphics[width=\textwidth]{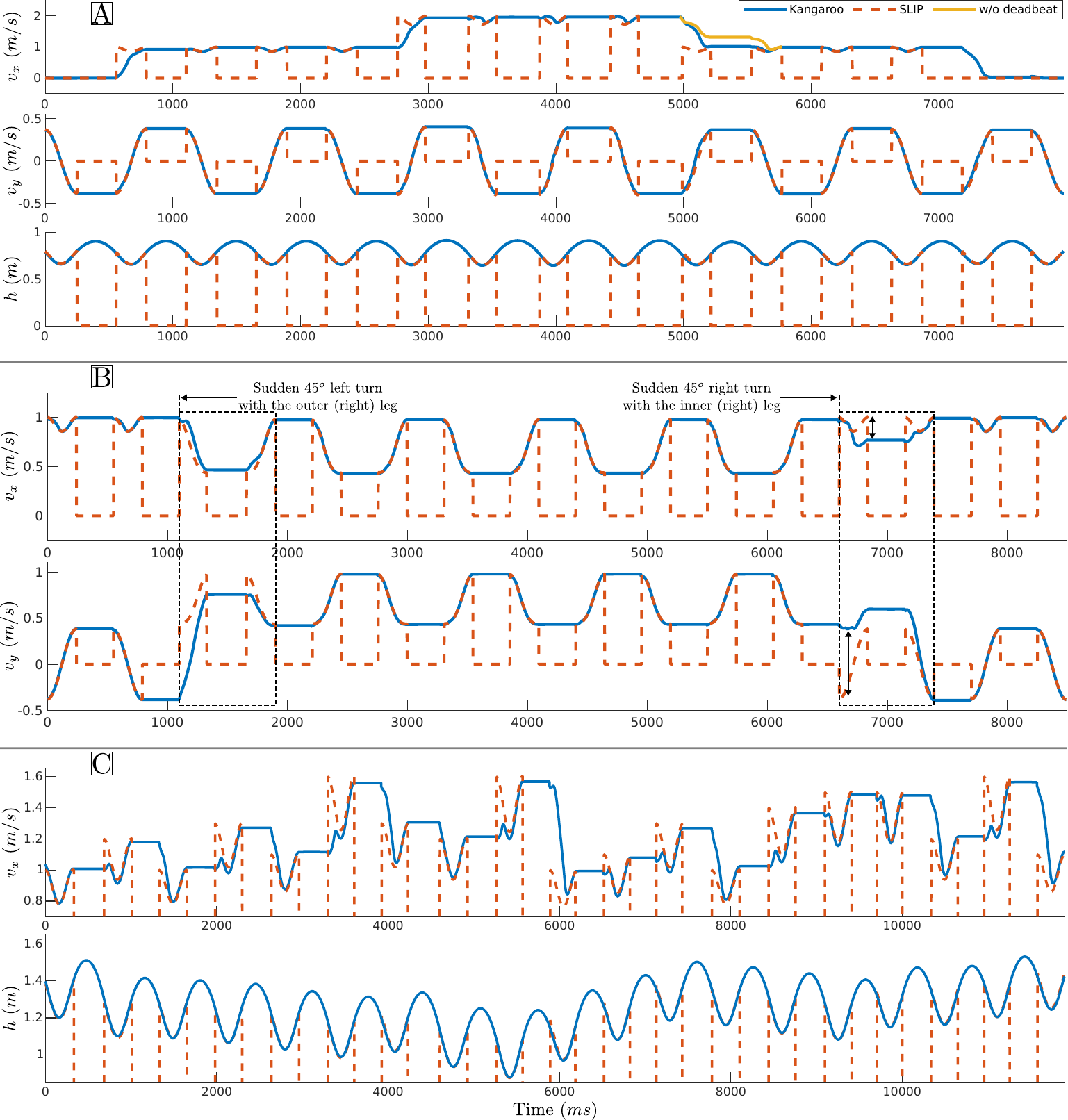}}
\caption{Response of Kangaroo to (A) sudden velocity command changes, (B) sudden direction changes by $\pi/4$ both with the inner and outer leg, (C) running through randomly generated stepping stones. Zero SLIP values indicate the flight phases.}
\label{combinedFigures}
\end{figure*}
\subsection{Response to Sudden Velocity Command Changes}
One of the key factors in navigating a random environment is the system's ability to rapidly switch trajectories. We plot the system's response to suddenly changing velocity commands in Fig.~\ref{combinedFigures}A. The figure illustrates that when commanded at the liftoff moment, the robot transitions from stationary jogging to a $1m/s$ running velocity in a single step. The robot then accelerates to $2m/s$, decelerates back to $1m/s$, and comes to a sudden stop. During the flight phase, in the event of an error or a trajectory change command, the deadbeat controller adjusts the leg angle and length parameters to ensure that the required ground reaction forces necessary to converge on a new trajectory or reject disturbances are feasible. The figure also compares the system's response to a sudden deceleration of $2\rightarrow1 m/s$ without deadbeat control. It shows that without any leg adjustment, the stance phase active control system cannot converge to the desired trajectory in a single step due to feasibility limitations. This is most important for running with precise foot placement point control and shows the significance of mapping SLIP input parameters with the deadbeat control gains in the step adaptation policy. Without deadbeat control, the response against disturbances and sudden velocity/direction commands usually ended up with failures in our simulations.
\subsection{Response to Sudden Direction Command Changes}
The deadbeat control implicitly inherits sudden and continuous direction change capabilities at the CoM trajectory generation level. The remaining inertial and posture control is achieved at the WBC level, also with the help of reactive limb-swing capabilities. The supplemental video shows that the robot can rotate $10^{o}$ per step and follow a sinusoidal (slalom) trajectory with $2m/s$ running velocity. We show the system's response to sudden $\pi/4$ radian direction change commands while running at $1m/s$ forward velocity, both with the inner and outer legs. At the touchdown, the spring-mass trajectories and the body rotation commands of the torso and foot are transformed first by $\pi/4$ radian for turning with the outer leg and then by $-\pi/4$ radian back for turning with the inner leg (see Fig.~\ref{combinedFigures}B). As the outer leg is already on the right side during the left turn, the feasible ground reaction forces match the required ones, and the system approaches the desired velocity states in a single stance. The second turn, on the other hand, is a right turn with the right leg. The robot's current posture and foot placement point do not allow it to converge to the desired states as much, and significant magnitude errors remain, especially in lateral velocity. The robot rolls over to the side and falls without any step adjustment in the next stance. The deadbeat controller places the foot wider than its periodic placement point such that the remaining lateral velocity is rejected. As a high-magnitude sudden turn also requires significant whole-body coordination to rotate the body, our simulations without reactive-limb swing control resulted in postural instability.
\subsection{Response to Randomly Generated Stepping Stones}
The combination of highly maneuverable velocity control, reactive limb swing control, and the step adaptation policy with convergence considerations enables the robot to run over randomly generated stepping stones, knowing only two steps ahead. The range of the randomness are $\{0.6,1.0\}m$ in longitudinal, $\{0.35,0.45\}m$ in lateral, and $\{-0.1,0.1\}m$ in vertical directions. The absolute distance between the two consecutive stepping stones is usually longer than the leg length, and due to randomness, each simulation run generates a different environment. The response to randomly generated stepping stones is shown in Fig.~\ref{combinedFigures}C. The figure illustrates the capability of making frequent trajectory changes at each step.
\subsection{Ground Reaction Forces}
\begin{figure}[h!]
\centerline{\includegraphics[width=0.95\columnwidth]{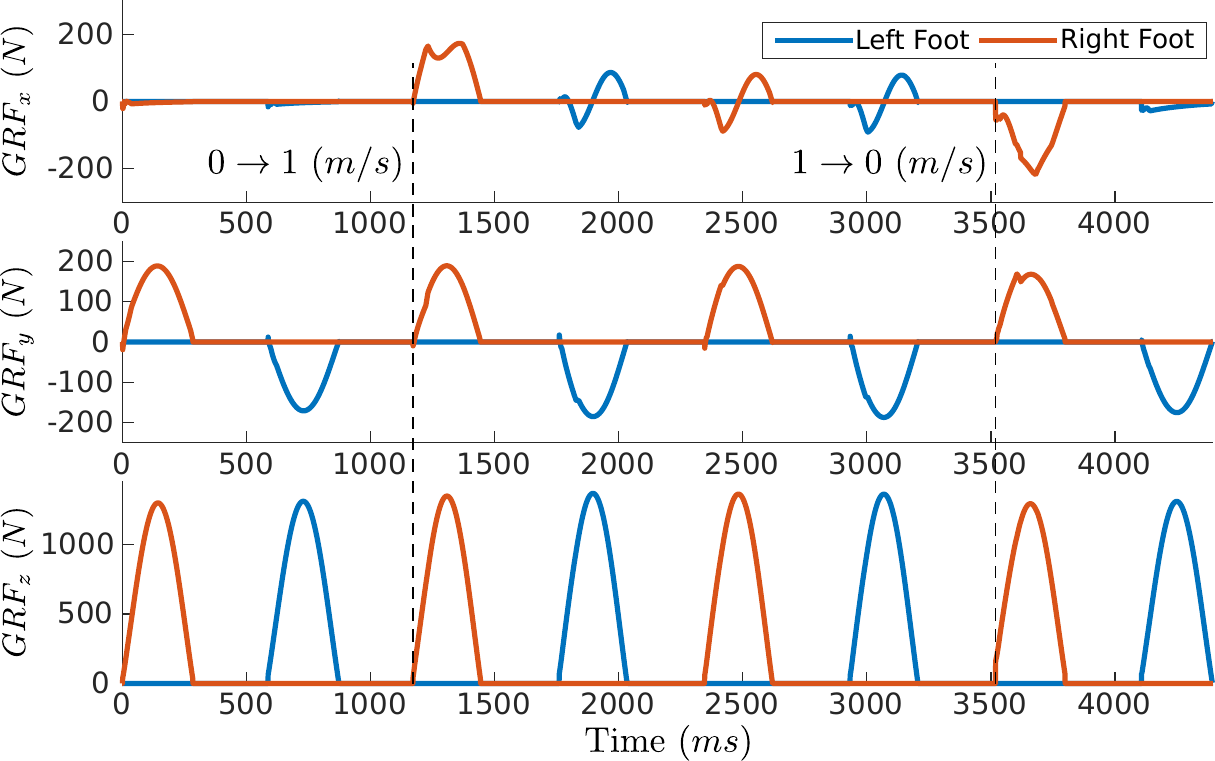}}
\caption{Ground reaction forces of Kangaroo while running at different paces.}
\label{groundReactionFig}
\end{figure}

Observing the system's ground reaction force evolution is another way to examine the match between the whole body and spring-mass dynamics. As shown in Fig.~\ref{groundReactionFig}, the ground reaction profiles of the humanoid running resemble those of a spring-mass dynamics system. As the desired running velocity changes, the active controller breaks the periodicity to converge into the desired trajectory. Once converged, the ground reaction forces follow almost the exact pattern of the conventional 3D spring-mass model.
\subsection{Response to Noise and Uncertainty}
\begin{figure} [t!]
\centerline{\includegraphics[width=0.85\columnwidth]{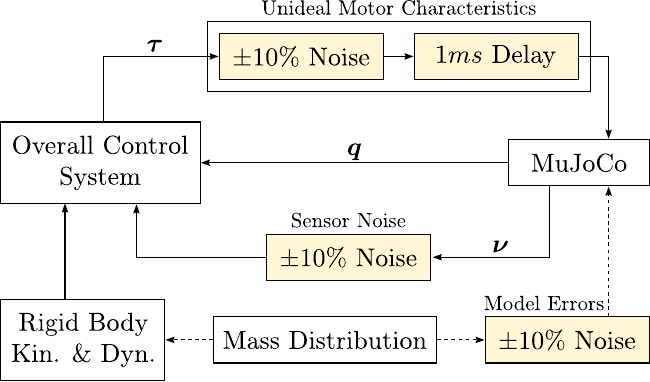}}
\caption{A diagram regarding the applied noise. The generalized velocity readings (body frame and joint velocities), motor torque commands, and mass distribution are subject to signal noise, delay, and uncertainty.}
\label{noiseDiagram}
\end{figure}

To better justify the proposed method's robustness against real-world challenges, such as disturbances, signal noises, imprecision, and delays, we perform an additional verification. The supplemental video already includes the system's response to force disturbances, undetected height changes, and corner-stepping conditions. Here, we take into account unideal actuation characteristics, delays, modeling errors, and sensor noise. The new system diagram, incorporating noise and uncertainty, is shown in Fig.~\ref{noiseDiagram}. The applied noise and uncertainties are random and generated in run-time through ``\textit{std::random\_device}'' and ``\textit{std::mt19937}'' in C++.

\begin{figure} [b!]
\centerline{\includegraphics[width=\columnwidth]{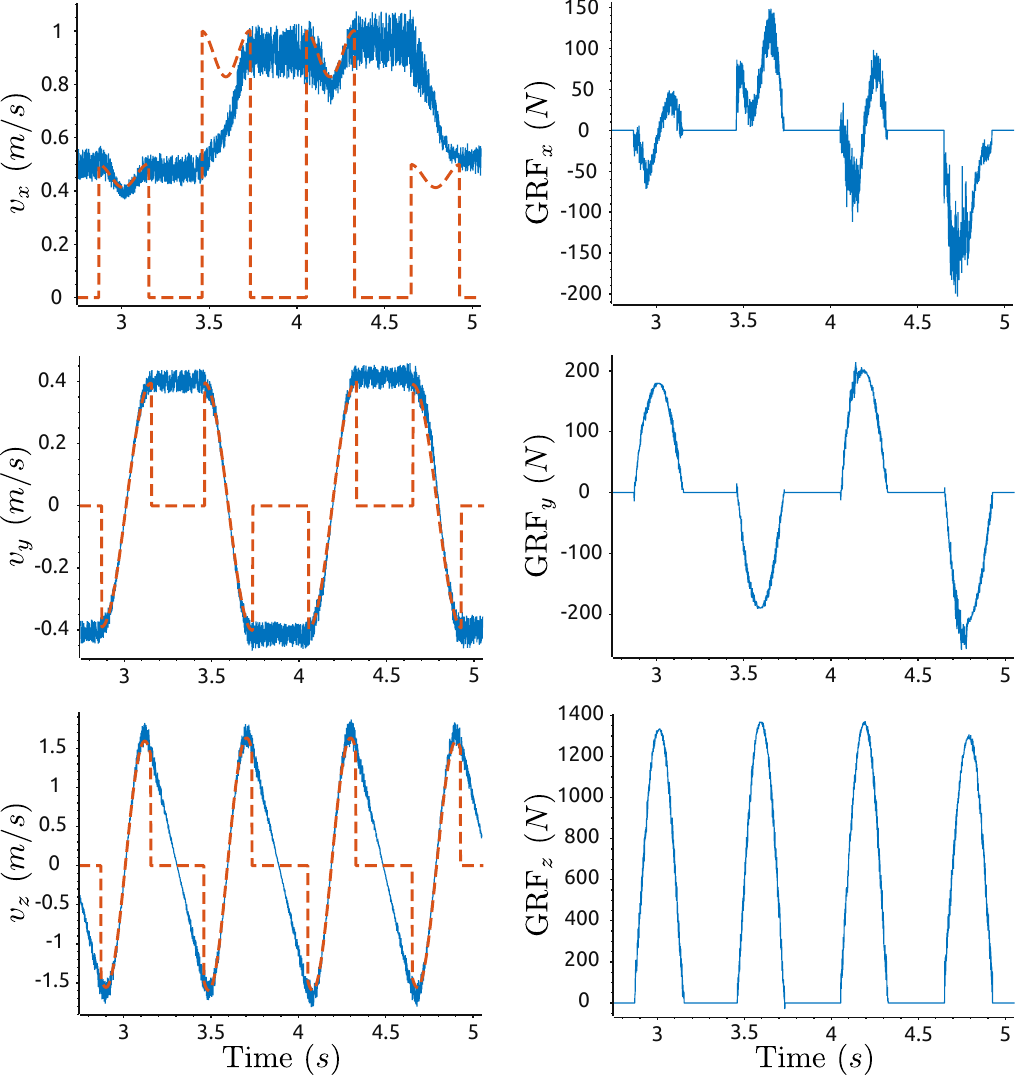}}
\caption{Kangaroo's center of mass velocities (left) and estimated ground reaction forces (right) under the influence of noise described in Fig.~\ref{noiseDiagram}.}
\label{noiseFig}
\end{figure}

We again examine running at different velocities and during transitions under the influence of the newly introduced real-world conditions. Note that the overall control system, in terms of the whole-body control, trajectories, and control gains, has been kept the same, and no additional tuning has been done. The system's response is provided in Fig.~\ref{noiseFig}. Compared to the ideal behaviors shown in Fig.~\ref{combinedFigures}, Fig.~\ref{noiseFig} exhibits a noisier behavior, which is more pronounced in the running direction, $v_x$. The actual periodic running velocity varies between the consecutive steps, and transitioning performance decays under the given conditions. Nevertheless, longitudinal velocity $\text{RMSE} = 0.085m/s$, lateral velocity $\text{RMSE} = 0.021m/s$, confirming stability under disturbances. Furthermore, state-based phase transitioning of the spring-mass model enhances the disturbance rejection performance as the flight and stance times are not precise under disturbances and uncertainties.
\section{Discussion}
Two main outcomes of this paper are step adaptation through the spring-mass model and reactive limb swing control.
\subsection{Step Adaptation}
The original spring-mass model is a velocity-based biomimetic running model constructed around periodic trajectories \cite{slipref1, wensingRunning, sovukluk_slip}. Although it exhibits highly dynamic behaviors over a compelling set of continuous ground reaction force patterns, step adaptation has not yet been achieved on a humanoid model. We demonstrate that, while retaining the advantages of the periodic spring-mass running model, our method can induce sudden changes in velocity (see Figures~\ref{combinedFigures}A-\ref{groundReactionFig}) and reject disturbances including physical perturbations, model and sensing imprecisions, and varying surface conditions (see Figures~\ref{firstPageFig},\ref{noiseDiagram}, and \ref{noiseFig}). The rapid trajectory switching capability is employed by the step adaptation policy (see Algorithms~\ref{alg1}-\ref{alg4}), enabling step adaptation in random environments (see Figures~\ref{flatRunFig}-\ref{combinedFigures}C). Although requiring a trajectory library, the library generation is entirely automatic and tuning-free, taking only a few seconds to calculate for hundreds of trajectories. It is scalable up to thousands of trajectories while still satisfying real-time constraints (see Table~\ref{timeTab}).
\subsection{Reactive Limb Swing Control}
The spring-mass model falls short in capturing multibody and angular momentum effects when controlling a full-scale humanoid robot \cite{wensingRunning, sovukluk_slip, sovukluk2025_icra}. As it only covers the CoM control and the foot placement point, the remaining limb swing and posture control are not considered in planning and must be handled separately. The limb swing is important both for angular momentum control and preparation for the next step. In cases of frequent trajectory switching and disturbances, where force profiles, stance time, flight time, and foot placement conditions vary, a different limb-swing behavior is required to maintain postural stability. We address this problem through a reactive limb-swing task, eliminating the need for trajectory-specific tuning. We demonstrate that the reactive limb swing task utilizes a backward leg swing for angular momentum compensation in the case of the Kangaroo (without arms) and employs both leg and arm swing in the case of the Unitree G1, thereby validating the framework's adaptability to kinematic variability. Along with tuning-free CoM trajectory switching capabilities, the limb swing task also enhances disturbance rejection performance at the whole-body control level, as it is related to overall postural stabilization while tracking the given CoM trajectory.
\section{Conclusion}
This study develops an overall control framework for highly maneuverable robots to navigate through random environments. First, a carefree 3D spring-mass trajectory library is generated by varying the running velocity, apex height, and leg stiffness. Then, an apex-to-apex deadbeat control gain library, which manipulates leg angles and length parameters for disturbance rejection or trajectory switching purposes, is obtained through a model that resembles the full-order model's constrained and actively controlled CoM dynamics well. The total offline library calculation process takes around $4.5$ seconds for $315$ trajectories and can even be done a few seconds before running the robot. Then, we develop a step adaptation policy that enables the robot to run in random environments. We show that running through random stepping stones and jumping over obstacles is possible with at least two-step and one-step planning, respectively. Furthermore, the step adaptation is performed considering the kinematic and dynamic feasibility, as well as convergence conditions. The online step-adaptation policy can handle thousands of trajectories while still satisfying the real-time conditions. Mapping the spring-mass trajectories to a full-size humanoid robot model is achieved through a conventional whole-body control with additional reactive-limb swing control formulation. The reactive limb swing enables the preservation of postural whole-body stability under a wide variety of different spring-mass trajectories and enhances the disturbance rejection performance. We show the inclusiveness and the robustness of the proposed framework through various challenging and agile behaviors, such as running through randomly generated stepping stones, jumping over random obstacles, performing highly dynamic slalom motions, changing the running direction suddenly with a random leg, and rejecting significant disturbances and uncertainties through the MuJoCo physics simulator. To further justify robustness against real-world challenges, we provide additional simulations with unideal actuation characteristics, delays, modeling errors, and sensor noise. All the aforementioned behaviors are performed using a single library and the same set of whole-body control parameters, without requiring additional tuning.
\section{Supplemental Media}
The presented and additional results are provided in the supplemental video with the following content:
\begin{itemize}
    \itemsep0em
    \item The actively controlled template model runs through randomly generated stepping stones.
    \item Starting from stationary jogging, Unitree G1 accelerates to $1m/s$ and comes back to stop.
    \item Starting from stationary jogging, Kangaroo first accelerates to $1m/s$, then to $2m/s$, and comes back to stop.
    \item Kangaroo runs at $2m/s$ velocity and performs slalom movement with $10$ degree rotation per step. Then it performs sudden $\pm \pi/4$ radian turns with random legs.
    \item Kangaroo runs at $1m/s$ velocity and jumps over randomly generated objects.
    \item Kangaroo runs through randomly generated stepping stones.
    \item Force disturbance rejection performance in all directions while running at $1m/s$ velocity.
    \item Running through unobserved ground height differences and edge stepping.
\end{itemize}
\section*{Acknowledgements}
This project has received funding from the European Research Council (ERC) under the European Union’s Horizon 2020 research and innovation programme (grant agreement No. 819358).
\bibliographystyle{elsarticle-num} 
\bibliography{references}
%
%
\end{document}